\documentclass[10pt,journal,compsoc]{IEEEtran}
\usepackage{amsmath,amsfonts}
\usepackage{algorithm}
\usepackage{algorithmic}
\usepackage{array}
\usepackage[caption=false,font=normalsize,labelfont=sf,textfont=sf]{subfig}
\usepackage{textcomp}
\usepackage{stfloats}
\usepackage{url}
\usepackage{verbatim}
\usepackage{cite}
\usepackage{afterpage}
\usepackage{floatpag}
\usepackage{ragged2e}
\usepackage{pifont}
\newcommand{\cmark}{\ding{51}}
\newcommand{\xmark}{\ding{55}}

\usepackage{epigraph}
\newcommand{\reffig}[1]{{Fig.~\ref{fig:#1}}}
\newcommand{\reftab}[1]{{Table~\ref{tab:#1}}}
\newcommand{\refsec}[1]{{Sec.~\ref{sec:#1}}}
\newcommand{\refeq}[1]{{Eq.~\eqref{eq:#1}}}
\newcommand{\refalg}[1]{{Algorithm~\ref{alg:#1}}}
\usepackage{booktabs}
\usepackage{multirow}
\usepackage[normalem]{ulem}
\usepackage{bm}
\usepackage{colortbl}

\newcommand{\revise}[1]{{\color{black}#1}}
\newcommand{\revisetwo}[1]{{\color{black}#1}}

\usepackage{tikz}
\usepackage{pgfplots}
\pgfplotsset{compat=1.18}

\usepackage[pagebackref,breaklinks,colorlinks]{hyperref}

\begin{document}\sloppy

\title{Lagrangian Motion Fields for \\Long-term Motion Generation}

% \author{
%     Yifei Yang,
%     Zikai Huang,
%     Chenshu Xu,
%     Shengfeng He,~\IEEEmembership{Member,~IEEE}
%     % <-this % stops a space
%     \thanks{Yifei Yang is with the Department of Computer Science, Singapore Management University, Singapore. (e-mail: yangyfaker@gmail.com).}
%     \thanks{Zikai Huang is with the School of Computer Science and Engineering, South China University of Technology, China. (e-mail: 202210188523@mail.scut.edu.cn).}
%     \thanks{Chenshu Xu is with the Department of Computer Science, Singapore Management University, Singapore. (e-mail: csxzxcs@gmail.com).}
%     \thanks{Shengfeng He is with the Department of Computer Science, Singapore Management University, Singapore. (e-mail: shengfenghe7@gmail.com).}
% }
\author{Yifei~Yang,
        Zikai~Huang,
        Chenshu~Xu,
        and~Shengfeng~He,~\IEEEmembership{Senior Member,~IEEE}
    \IEEEcompsocitemizethanks{
    \IEEEcompsocthanksitem This work is supported by the Guangdong Natural Science Funds for Distinguished Young Scholars (Grant 2023B1515020097), the Singapore Ministry of Education AcRF Tier 1 Grant (Grant No.: MSS25C004), and the Lee Kong Chian Fellowships. Yifei Yang and Zikai Huang contributed equally to this work. Corresponding author: Shengfeng He.
    \IEEEcompsocthanksitem Yifei Yang, Chenshu Xu, and Shengfeng He are with the School of Computing and Information Systems, Singapore Management University, Singapore. E-mail: yangyfaker@gmail.com; csxzxcs@gmail.com; shengfenghe@smu.edu.sg.
    \IEEEcompsocthanksitem Zikai Huang is with the School of Computer Science and Engineering, South China University of Technology, China. E-mail: 202210188523@mail.scut.edu.cn.
    }
}

% The paper headers
\markboth{IEEE Transactions on Pattern Analysis and Machine Intelligence}%
{Yang \MakeLowercase{\textit{et al.}}: Lagrangian Motion Fields for Long-term Motion Generation}

\IEEEtitleabstractindextext{\justify
% TODO: under construction
\begin{abstract}
Long-term motion generation is a challenging task that requires producing coherent and realistic sequences over extended durations. Current methods primarily rely on framewise motion representations, which capture only static spatial details and overlook temporal dynamics. This approach leads to significant redundancy across the temporal dimension, complicating the generation of effective long-term motion. To overcome these limitations, we introduce the novel concept of Lagrangian Motion Fields, specifically designed for long-term motion generation. 
By treating each joint as a Lagrangian particle with uniform velocity over short intervals, our approach condenses motion representations into a series of ``supermotions'' (analogous to superpixels). This method seamlessly integrates static spatial information with interpretable temporal dynamics, transcending the limitations of existing network architectures and motion sequence content types. Our solution is versatile and lightweight, eliminating the need for neural network preprocessing.
Our approach excels in tasks such as long-term music-to-dance generation and text-to-motion generation, offering enhanced efficiency, superior generation quality, and greater diversity compared to existing methods. Additionally, the adaptability of Lagrangian Motion Fields extends to applications like infinite motion looping and fine-grained controlled motion generation, highlighting its broad utility. Video demonstrations are available at \url{https://plyfager.github.io/LaMoG}.
\end{abstract} 

\begin{IEEEkeywords}
Motion Generation, Animation, Motion Representations
\end{IEEEkeywords}
}
% \IEEEpubid{0000--0000/00\$00.00~\copyright~2021 IEEE}
% Remember, if you use this you must call \IEEEpubidadjcol in the second
% column for its text to clear the IEEEpubid mark.

\maketitle
\begin{figure*}[t] % [t] 确保图像放在页面顶部，使用 figure* 环境跨双栏
    \centering
    \includegraphics[width=.85\linewidth]{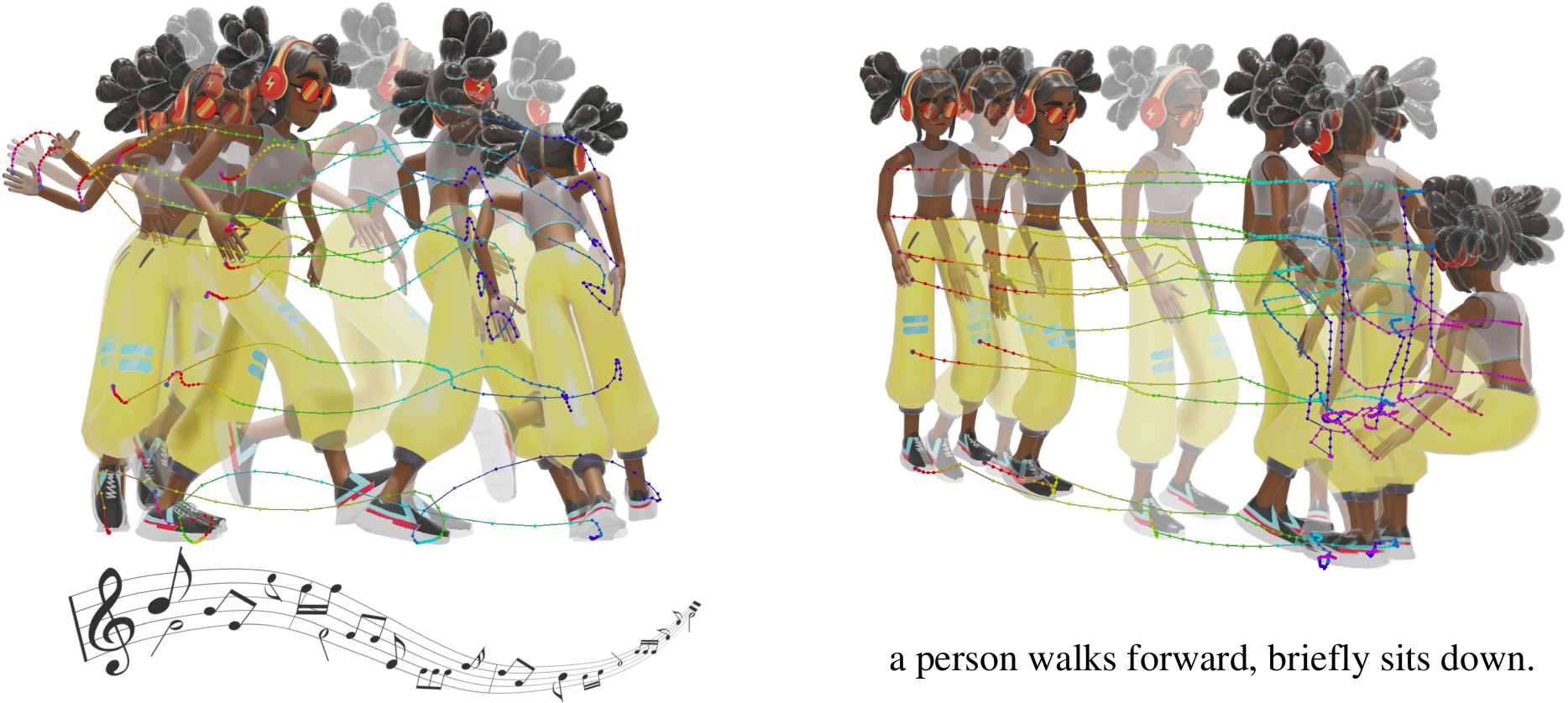}
    \caption{We propose a novel motion representation, Lagrangian Motion Fields, for long-term motion generation. This method generates abstract motion segments called ``supermotions'', enabling the accelerated production of high-quality long-term motions, such as dance generation (left) and text-to-motion (right). The trajectories illustrate the motion paths of each joint, with distinct colors representing different supermotion segments. Each supermotion segment comprises an initial pose (alternating between opaque and transparent characters for clarity), a motion field (represented by the trajectories of various joints in the same color), and a time duration (indicated by the length of the segments in the same color).}
    \label{fig:teaser}
\end{figure*}
% \begin{abstract}
% This document describes the most common article elements and how to use the IEEEtran class with \LaTeX \ to produce files that are suitable for submission to the IEEE.  IEEEtran can produce conference, journal, and technical note (correspondence) papers with a suitable choice of class options. 
% \end{abstract}

\section{Introduction}
\setlength\epigraphwidth{8cm}
\setlength\epigraphrule{0pt}
\epigraph{\emph{``Be water, my friend.''}}{--- {Bruce Lee}}
\IEEEPARstart{L}{ong}-term 3D human motion generation is crucial in fields such as computer animation, virtual reality, and human-computer interaction, as it enables the creation of authentic, dynamic movements over extended periods. Realistic motion sequences greatly enhance immersion, making virtual environments more believable and engaging. This level of realism is vital for applications in gaming, film, therapy, sports training, and remote communication, where it significantly improves user experience and effectiveness.

Previous approaches to long-term 3D human motion generation have primarily addressed the challenge from two perspectives: generation paradigms and motion representations, each with inherent constraints and limitations.
From the generation paradigm perspective, autoregressive methods~\cite{fact, ao2023gesturediffuclip} theoretically offer the flexibility to generate motion sequences of any length. However, in practice, these methods often suffer from error accumulation, resulting in unrealistic or stagnant motion sequences~\cite{zhuang2022music2dance, bailando, lodge, sun2022you}. Additionally, generating long sequences with autoregressive methods requires repeated inference, leading to inefficiencies and necessitating trade-offs between sequence length, quality, and computational complexity.

Recently, diffusion models~\cite{ddpm, ddim} have made significant strides in 3D motion generation. For example, EDGE~\cite{edge} introduced a parallel generation strategy, producing long sequences in batches with overlapping segments. Subsequent works~\cite{lodge, priormdm} have focused on refining transitions between segments to achieve more natural and harmonious sequences. Despite these advancements, the process of stitching together motion segments can lead to fragmented global coherence, resulting in motion sequences that lack fluidity and appear unrealistic. Moreover, these methods incur high computational overhead, as they require generating each frame individually, leading to inefficiencies.

In contrast to previous works, we believe that the key to effective long-sequence motion generation lies in adopting a more efficient motion representation scheme that balances compactness, robust generalization, and interpretability. Instead of predicting spatial coordinates for each joint frame by frame, we propose treating human motion as a dynamic flow over time to explicitly capture evolving trends and better understand its dynamic characteristics.

Specifically, we introduce Lagrangian Motion Fields, a concept inspired by fluid dynamics, where each joint is treated as a Lagrangian particle with uniform velocity over short time intervals. This approach provides a unique perspective for modeling flow-like 3D human motion. Recognizing that complex human movements can often be decomposed into simpler, uniform motions, we also draw inspiration from superpixels~\cite{supercnn, achanta2012slic} in 2D image processing. Our Lagrangian Motion Fields generate abstract motion segments, termed ``supermotions'', resulting in a more compact motion representation.

Building on the concept of Lagrangian Motion Fields, we introduce a general two-stage generation pipeline that can be seamlessly integrated into any motion generation network. In the first stage, we generate a supermotion sequence, which is then decompressed into the full-resolution motion sequence using Lagrangian Motion Fields. The reduced temporal resolution of the supermotion sequence makes it more compact and efficient to generate, enabling the creation of long-term motion sequences with minimal computational overhead. Furthermore, the supermotion representation supports diverse applications, including infinite motion looping and duration-controlled motion generation. To further refine and diversify the generated supermotion sequence, we introduce a lightweight motion refinement network designed to enhance the global coherence of the motion. Extensive experiments demonstrate the effectiveness of our Lagrangian Motion Fields in two downstream applications: text-to-motion and dance generation, as illustrated in Fig.~\ref{fig:teaser}.

Our contributions are summarized as follows:
\begin{itemize}\nointerlineskip
    \item We introduce Lagrangian Motion Fields for 3D human motion generation, leading to the creation of the abstract motion segment ``supermotion''. This approach significantly simplifies temporal representation, ensuring computational efficiency without the need for network extraction. Additionally, it is generalizable across different motion data types and maintains intuitive physical interpretability.
    \item We propose a general pipeline for long-term motion generation that utilizes the supermotion representation and can be seamlessly integrated into any motion generation network.
    \item Extensive experiments demonstrate the effectiveness and generalizability of our method in long-term motion generation tasks, outperforming state-of-the-art methods in both music-to-dance and text-to-motion generation.
    \item The proposed Lagrangian Motion Fields enable a variety of applications, including infinite motion looping and duration-controlled motion generation.
\end{itemize}
\section{Related Works}
\textbf{Motion Representation. }
Existing methods for 3D motion generation can be broadly classified into two categories: framewise representations and compressed representations.

Framewise representations, such as rotation matrices or Cartesian coordinates, are commonly used in motion generation tasks\cite{mdm, edge, lodge, tm2t, flame, oohmg, teach, fact, transflower, priormdm, mofusion}. While intuitive, these representations suffer from high redundancy and lack temporal information, as each frame is represented independently, failing to capture the temporal coherence of the motion sequence explicitly.
In contrast, methods employing VQ-VAE~\cite{dancemeld, bailando, bailando++, t2m-gpt, yi2023generating, humantomato, tm2t, ude} compress motion sequences into discrete codes that can be decoded to reconstruct the original sequences. However, these approaches have limited generalization capabilities, often requiring retraining of the codebook for different datasets. Additionally, the codes in the codebook are difficult to interpret, limiting their applicability and posing challenges for extending them to other applications, such as infinite motion looping.

In the realm of 2D image processing, superpixels denote clusters of spatially contiguous pixels sharing similar attributes, such as color or intensity. Previous research~\cite{supercnn, achanta2012slic, jampani2018superpixel, gadde2016superpixel, xu2022groupvit, zhang2023clip} has leveraged this representation to reduce image data complexity, making it more manageable for processing and analysis while preserving essential information and structures.
Drawing inspiration from these methods, we propose supermotion as a compact representation of extended motion sequences, capturing essential motion characteristics. By incorporating Lagrangian Motion Fields, the proposed supermotion representation captures both static spatial information and interpretable temporal dynamics.

\noindent\textbf{Long-term Motion Generation. }
Long-term motion generation poses a long-standing challenge in motion generation research. Autoregressive methods~\cite{fact, transflower, wen2021autoregressive, zhang2024bidirectional, ao2023gesturediffuclip, como} theoretically offer the capacity to produce motion sequences of arbitrary length. However, they frequently encounter the notorious issue of error accumulation, resulting in unrealistic or frozen motion.

Diffusion models~\cite{ddpm, ddim, glide, dalle, diffdance, lda} have demonstrated continuous breakthroughs across various domains. MDM~\cite{mdm} marks a pioneering effort in applying diffusion-based techniques to motion generation, showcasing their efficacy. Notably, diffusion models exhibit enhanced generative capabilities compared to alternative network types, along with robust zero-shot inpainting abilities. Subsequent works~\cite{edge, priormdm, lodge, zhang2023diffcollage, longdancediff} have refined motion generation processes by modifying the diffusion sampling paradigm and improving transitions between segments. For instance, EDGE~\cite{edge} proposed a parallel generation strategy, where each denoising step directly utilizes the latter half of the preceding segment as the former half of the subsequent segment, subsequently unfolding them into a complete motion sequence. Building upon this, PriorMDM~\cite{priormdm} introduced DoubleTake, refining transitions in two stages: the first stage uses weighted blending based on EDGE, and the second employs soft-masking and linear-masking inpainting for refinement.

While these methods have enhanced the quality of long-sequence generation to some extent, they still consider only a finite context window during generation, accounting for adjacent segments rather than the entire motion sequence comprehensively. Lodge~\cite{lodge} proposed the concept of dance primitives, utilizing choreography prior knowledge, and introduced a global-local framework to generate long-term dance motion sequences. Despite the ability of these methods to generate long-term motion sequences, they suffer from high redundancy and computational overhead due to their framewise motion representation.

To tackle these challenges, we propose Lagrangian Motion Fields, which simplifies the generation process by drastically reducing temporal resolution from the source. This approach offers a general motion representation applicable across various motion content types and can be integrated into any motion generation network.

\section{Method}
\begin{figure}
    \centering
    \includegraphics[width=0.49\textwidth]{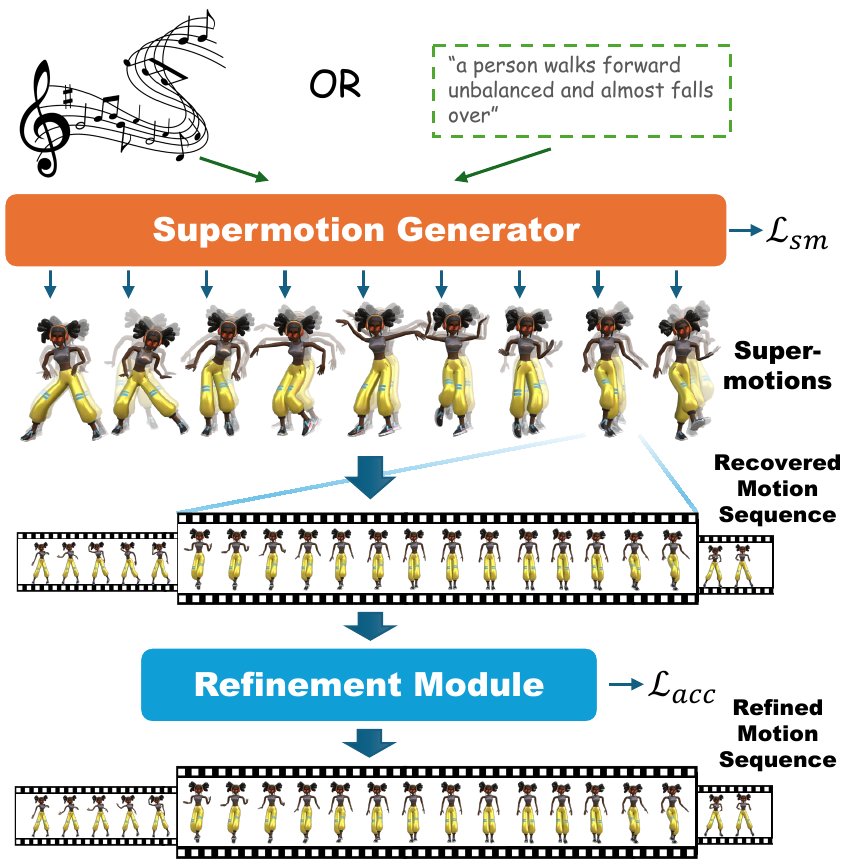}
    \caption{We propose a two-stage long-term motion generation pipeline. Given a control signal, we first generate the supermotion sequence, which is then recovered into a full-resolution motion sequence. The recovered motion sequence is further refined and diversified through a refinement module.}
    %\vspace{-4mm}
    \label{fig:pipeline}
\end{figure}

Our goal is to generate a long-term 3D human motion sequence $\mathbf{X} = \{\mathbf{x}_0, \mathbf{x}_1, \dots, \mathbf{x}_{N-1}\}$ with a given control signal $\mathbf{C}$. Here, $\mathbf{x}_i$ represents the generalized coordinates of joints at frame $i$, and $J$ denotes the number of joints. 
{We first present the Lagrangian motion fields for 3D human motion generation, as detailed in~\refsec{lagrangian_motion_fields}. In~\refsec{supermotion}, we introduce the supermotion representation, which compresses the original motion sequence into a series of supermotions.
The supermotion generation module is described in~\refsec{supermotion_generation_module}, which produces supermotions from the control signal. In~\refsec{refinement_module}, we detail the refinement module, which further refines the generated supermotions into the final motion sequence. 
Finally, in~\refsec{adaptation}, we discuss the adaptation of our method to downstream tasks. }

\subsection{Lagrangian Motion Fields}  \label{sec:lagrangian_motion_fields}
\revise{Previous work either treats motion as isolated spatial positions, neglecting explicit temporal domain information~\cite{edge, lodge, mdm}, or relies on latent-based representations~\cite{bailando, humantomato, ude}, which are difficult to interpret and generalize. On the other hand, human motion, much like fluid flow, follows continuous trajectories and exhibits coordinated dynamics that evolve smoothly over time. This fundamental similarity to fluid dynamics provides a theoretical basis for employing field-based modeling in human movement analysis. Since traditional Lagrangian fields in fluid mechanics are continuous, our approach adapts this concept by employing a discretized approximation for computational efficiency, analogous to finite element methods in computational fluid dynamics. Specifically, human motion maintains consistent semantic meaning over short time intervals, a property rooted in biomechanics. From a biomechanical perspective, this consistency arises from the smooth velocity transitions and continuous trajectories of individual joints. Consequently, modeling human motion as a continuous velocity field naturally captures both its spatial structure and temporal evolution.
To effectively model the temporal continuity of human motion, we propose Lagrangian motion fields for 3D human motion generation.
}

The Lagrangian perspective in fluid dynamics follows individual particles as they move through space and time, rather than relying on fixed spatial reference points, as in the Eulerian framework. Given the initial position $x_0$ of a particle, its position at any later time $t$ is determined by the Lagrangian mapping:
\begin{equation} \label{eq:lag_eqn}
x(t) = \mathbf{L}(x_0, t),
\end{equation}
where $\mathbf{L}$ is a function that maps the initial position and time to the current position.
This approach excels in capturing complex dynamics like mixing and turbulence~\cite{sommerfeld2017numerical, chrigui2005eulerian, luo2009multiscale}.

Building on this concept, we analogously model each joint as a particle whose trajectory is determined by time and control signals from an initial position, under the assumption that joint positions and velocities remain continuous, without abrupt changes or discontinuities.
\revise{Given each joint $j$ as a Lagrangian particle, the velocity field is defined as:
\begin{equation} \label{eq:lag_vel_field}
  {v}(\mathbf{X}_j, t) = \frac{\mathbf{x}_j(t + \Delta t) - \mathbf{x}_j(t)}{\Delta t}, \quad t \in [t_s, t_s + d_s],
\end{equation}
where $t_s$ denotes the start time of a motion segment $s$, and $d_s$ is its duration. This constitutes a discrete velocity field defined over joints and time segments. Within each motion segment $s$, the velocity $v(\cdot)$ serves as a local field, governing the motion of all joints during $t \in [t_s, t_s + d_s]$.}
% Under these assumptions, the kinematical behavior of each joint over time can be similarly quantified and predicted using Lagrangian motion fields as follows:
% \begin{equation} \label{eq:lag_motion_eq}
%     x_i^j = \mathbf{L}(x_0^j, \mathbf{C}, i), i \in \{0, N-1\}, j \in \{1, 2, \cdots, J\},
% \end{equation}
% where $x_i^j$ is the representation of the joint $j$ at frame $i$, $\mathbf{C}$ is the control signal and $\mathbf{L}$ is the Lagrangian motion fields function.
\revise{
The segment duration $d_s$ functions as a  \textit{Kolmogorov scale}~\cite{pope2001turbulent}, smoothing fine-scale variations for computational tractability while preserving essential motion characteristics.
Using this discrete velocity field representation, the kinematical behavior of each joint over time can be quantified and predicted as follows:
\begin{equation} \label{eq:lag_motion_eq}
    x_i^j = \mathbf{L}(x_0^j, \mathbf{C}, i), i \in \{0, N-1\}, j \in \{1, 2, \cdots, J\},
\end{equation}
where ${x}_i^j$ is the representation of the joint $j$ at frame $i$, $\mathbf{C}$ is the control signal, and $\mathbf{L}$ is the Lagrangian motion fields function that integrates the velocity field to obtain joint positions.}

{Notably, the proposed pipeline is not constrained to a specific coordinate system and can be applied universally.}
Through this formulation, our key insight is to shift from predicting discrete static spatial information snapshots at each frame (i.e., the framewise representation) to modeling the continuous flow of motion.

Our approach leverages the observation that human motion often exhibits similar trends over short periods. By modeling the flow of motion, we reduce data redundancy and enhance computational efficiency. Furthermore, the Lagrangian motion field representation is dataset-agnostic and can be generalized across different types of motion. Compared to latent-based representations~\cite{bailando, humantomato, ude}, Lagrangian motion fields provide a more interpretable representation of human motion, making them applicable to tasks such as infinite motion looping and fine-grained controlled motion generation.

\subsection{Supermotion} \label{sec:supermotion}
\begin{figure}
    \centering
    \includegraphics[width=0.49\textwidth]{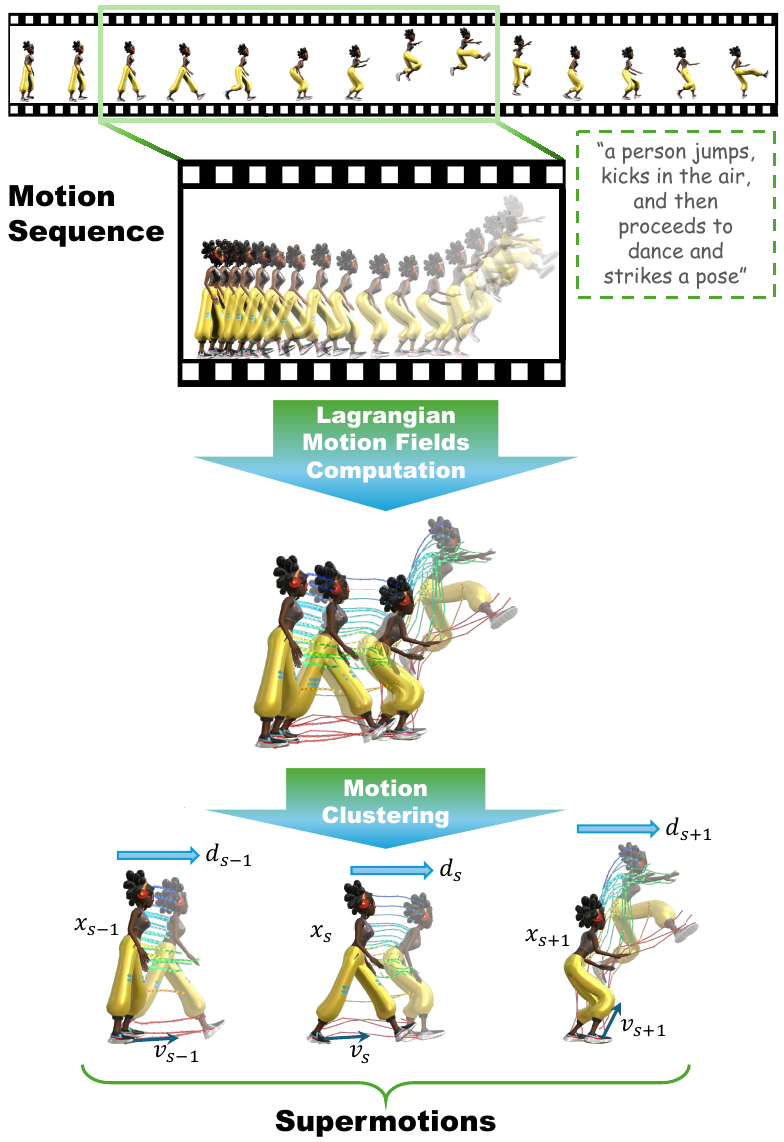}
    \caption{Given a full-resolution framewise motion sequence, we first compute the Lagrangian motion fields. We then cluster the sequence into segments approximated as uniform motion, each represented by a supermotion. }
    \label{fig:motion2supermotion}
\end{figure}

\begin{algorithm}
\caption{\revise{Motion to Supermotion Conversion}}
\label{alg:motion2super}
\revise{\begin{algorithmic}[1]
\STATE {\textbf{Input:}} Motion sequence set $\mathcal{M} = \{M_1, M_2, ..., M_N\}$, window size $w$, number of clusters $K$
\STATE {\textbf{Output:}} Supermotion sequences $\mathcal{S}$

\STATE // Phase 1: Compute velocities and assign labels
\FOR{each motion $M_i$ in $\mathcal{M}$}
    \STATE $V_i[t] \leftarrow M_i[t+1] - M_i[t], \forall t \in [0, L_i-1)$
\ENDFOR
\STATE $C \leftarrow \text{KMeans}(\{V_1,...,V_N\}, K)$
\STATE $l_i[t] \leftarrow C.\text{predict}(V_i[t]), \forall i \in [1,N], t \in [0,L_i-1)$

\STATE // Phase 2: Smooth labels and find segments
\FOR{each sequence of labels $l_i$}
    \STATE $\hat{l}_i[t] \leftarrow \mathbf{argmax_k} |\{l_i[j] = k : j \in [t-w/2, t+w/2]\}|$
    \STATE $P_i \leftarrow \{t : \hat{l}_i[t] \neq \hat{l}_i[t-1]\} \cup \{0,L_i\}$
\ENDFOR

\STATE // Phase 3: Generate supermotions
\FOR{each motion $M_i$ and points $P_i$}
    \STATE $S_i \leftarrow \emptyset$
    \FOR{each segment $[p_j, p_{j+1}]$ in $P_i$}
        \STATE $x_s \leftarrow M_i[p_j]$
        \STATE $v_s \leftarrow$ Mean($V_i[p_j:p_{j+1}]$)
        \STATE $d_s \leftarrow p_{j+1} - p_j$
        \STATE $S_i \leftarrow S_i \cup \{(x_s, v_s, d_s)\}$
    \ENDFOR
    \STATE $\mathcal{S} \leftarrow \mathcal{S} \cup \{S_i\}$
\ENDFOR

\RETURN {} $\mathcal{S}$
\end{algorithmic}}
\end{algorithm}

\revise{The primary challenge in long-term motion generation is efficiently representing extended motion sequences while preserving essential information. While Lagrangian motion fields provide a structured theoretical framework for capturing motion dynamics in a discrete form, a practical mechanism is needed to obtain meaningful discretized motion segments.  
Drawing inspiration from superpixel techniques in 2D image processing~\cite{supercnn, achanta2012slic, xu2022groupvit}, we introduce \textbf{supermotion} as a fundamental abstract motion unit within our Lagrangian framework. Supermotion segments motion into compact units with similar characteristics, enabling efficient compression while maintaining structural coherence. By encapsulating recurrent motion motifs, supermotion reduces the sequence's temporal resolution, simplifying the generation process while preserving essential dynamics.}

\revise{As illustrated in~\reffig{motion2supermotion} and detailed in~\refalg{motion2super},} for a lengthy motion sequence $\mathbf{X}$, we initially compute the Lagrangian motion flow across the entire sequence. 
{Subsequently, we apply a K-means clustering $\phi(\cdot)$ on the motion fields to classify each part of the motion sequence. 
After obtaining the motion labels, we smooth the labels and group the adjacent identical labels to form coherent segments $\{\mathbf{seg}_s\}_{s=0}^{M-1}$.
\begin{equation}
    % \tilde{\phi}(x_t) = \arg\max_{k \in \{1, \dots, K\}} \sum_{i=-w}^{w} \mathbb{I}(\phi(x_{t+i}) = k), \\
     % \{\mathbf{seg}_s\}_{s=0}^{M-1} =  \mathbf{Group}(\{\tilde{\phi}(x_t)\}_{t=0}^{N-1}),
     \{\mathbf{seg}_s\}_{s=0}^{M-1} =  \mathbf{Group}(\phi(\mathbf{X}, K)),
\end{equation}
where $K$ is the number of clusters, and $M \ll N$. 
% The operation $\mathbf{Group}$ is responsible for partitioning the label sequence into $M$ contiguous segments, denoted as $\mathbf{seg}_s$.
}
We define a supermotion $\mathbf{sm}_s = [\mathbf{x}_s, \mathbf{v}_s, d_s]$ to represent segment $\mathbf{seg}_s$, characterized by the starting position $\mathbf{x}_s$, the corresponding velocity $\mathbf{v}_s$, and the time duration $d_s$.
Notably, while segments may vary in length, the velocity remains relatively consistent within each segment.
For any given time $t$ within segment $\mathbf{seg}_s$, the position of a joint $x_t^j$ {can be recovered from the supermotion $\mathbf{sm}_s$ following the definition of Lagrangian motion fields in~\refeq{lag_motion_eq}:
\begin{equation} \label{eq:sm2m}
\begin{split}
    x_t^j &= \mathbf{L}(x_{t_s}^j, \mathbf{C}, t) \approx x_{t_s}^j + v_{t_s}^j (t - t_s), \\
    t \in &\{t_s, t_s + d_s - 1\},
\end{split}
\end{equation}
where $t_s = \sum_{i=0}^{s-1}{d_i}$ denotes the starting time of the $s$-th segment.}

\subsection{Supermotion Generation Module} \label{sec:supermotion_generation_module}
As illustrated in~\reffig{pipeline}, the control signal $\mathbf{C}$ is fed into the supermotion generation module, which outputs only $M$ supermotions instead of the lengthy motion sequence of $N$ frames. The compact supermotion representation eliminates the computational burden of processing the long-range dependencies between control signals and generation modality by significantly reducing temporal resolution.

Given the versatility of our proposed supermotion representation, which seamlessly integrates into any motion generation network, we explore its applicability across various models by employing both diffusion-based~\cite{edge, lodge} and non-diffusion-based~\cite{mdm} methods for supermotion generation.
During training, we define the  basic reconstruction loss $\mathcal{L}_{recon}$ as follows:
\begin{equation} \label{eq:L_recon}
    \mathcal{L}_{recon} = \frac{1}{M}\sum_{s=0}^{M-1}{ \lVert{\mathbf{sm}_{s} - \hat{\mathbf{sm}}_{s}}\rVert_2^2},
\end{equation}
where $\mathbf{sm}_{s}$ denotes the ground truth supermotion, $\hat{\mathbf{sm}}_{s}$ denotes the generated supermotion.
Alongside, we incorporate auxiliary losses as in previous works ~\cite{edge, mdm}:
\begin{equation} \label{eq:L_joint}
    \mathcal{L}_{joint} = \frac{1}{M}\sum_{s=0}^{M-1}{ \lVert{FK(\mathbf{x}_{t_s}) - FK(\hat{\mathbf{x}}_{t_s})}\rVert_2^2},
\end{equation}
\begin{equation} \label{eq:L_vel}
    \mathcal{L}_{vel} = \frac{1}{M}\sum_{s=0}^{M-1}{ 
    {\lVert{{\mathbf{x}_{t_s}}{^\prime} - {\hat{\mathbf{x}}_{t_s}}{^\prime}} \rVert_2^2},
}
\end{equation}
\begin{equation} \label{eq:L_contact}
    \mathcal{L}_{contact} = \frac{1}{M}\sum_{s=0}^{M-1}{ \lVert{{FK_{foot}(\hat{\mathbf{x}}_{t_s})}{^\prime} \cdot \hat{g}_{t_s}}\rVert_2^2},
\end{equation}
where $FK(\cdot)$ denotes the forward kinematics function that converts the rotation position representation into 3D points in Cartesian space, $\mathbf{x}_{t_s}$ represents the ground truth motion at time $t_s$, $\hat{\mathbf{x}}_{t_s}$ denotes the generated motion at time $t_s$, $\mathbf{x}_{t_s}^\prime$ denotes the velocity of the ground truth motion at time $t_s$, $\hat{\mathbf{x}}_{t_s}^\prime$ denotes the velocity of the generated motion at time $t_s$, and $\hat{g}_{t_s}$ denotes the predicted binary feet contact label at time $t_s$.

Additionally, to ensure the generated supermotions align with our continuity assumption and avoid abrupt position changes, we introduce a supermotion coherence loss, promoting smoother transitions between adjacent supermotions:
% \todo{coherent loss is different for vel and 6d representation}
\begin{equation} \label{eq:coherent_loss}
    \mathcal{L}_{coherent} = \frac{1}{M-1}\sum_{s=0}^{M-2}{ \lVert{\hat{\mathbf{x}}_{s+1} - (\hat{\mathbf{x}}_s + \hat{\mathbf{v}}_s \cdot \hat{d}_s)}\rVert_2^2},
\end{equation}
The total loss function for the supermotion generation module is defined as:
\begin{equation} \label{eq:L_sm}
    \begin{split}
    \mathcal{L}_{sm} &= \mathcal{L}_{recon} + \lambda_{joint} \cdot \mathcal{L}_{joint} + \lambda_{vel} \cdot \mathcal{L}_{vel} + \\
    &\lambda_{contact} \cdot \mathcal{L}_{contact} + \lambda_{coherent} \cdot \mathcal{L}_{coherent},
    \end{split}
\end{equation}
where $\lambda_{joint}$, $\lambda_{vel}$, $\lambda_{contact}$, $\lambda_{coherent}$ are hyperparameters to balance the loss terms.

\subsection{Refinement Module} \label{sec:refinement_module}
\begin{figure}
    \centering
    \includegraphics[width=0.49\textwidth]{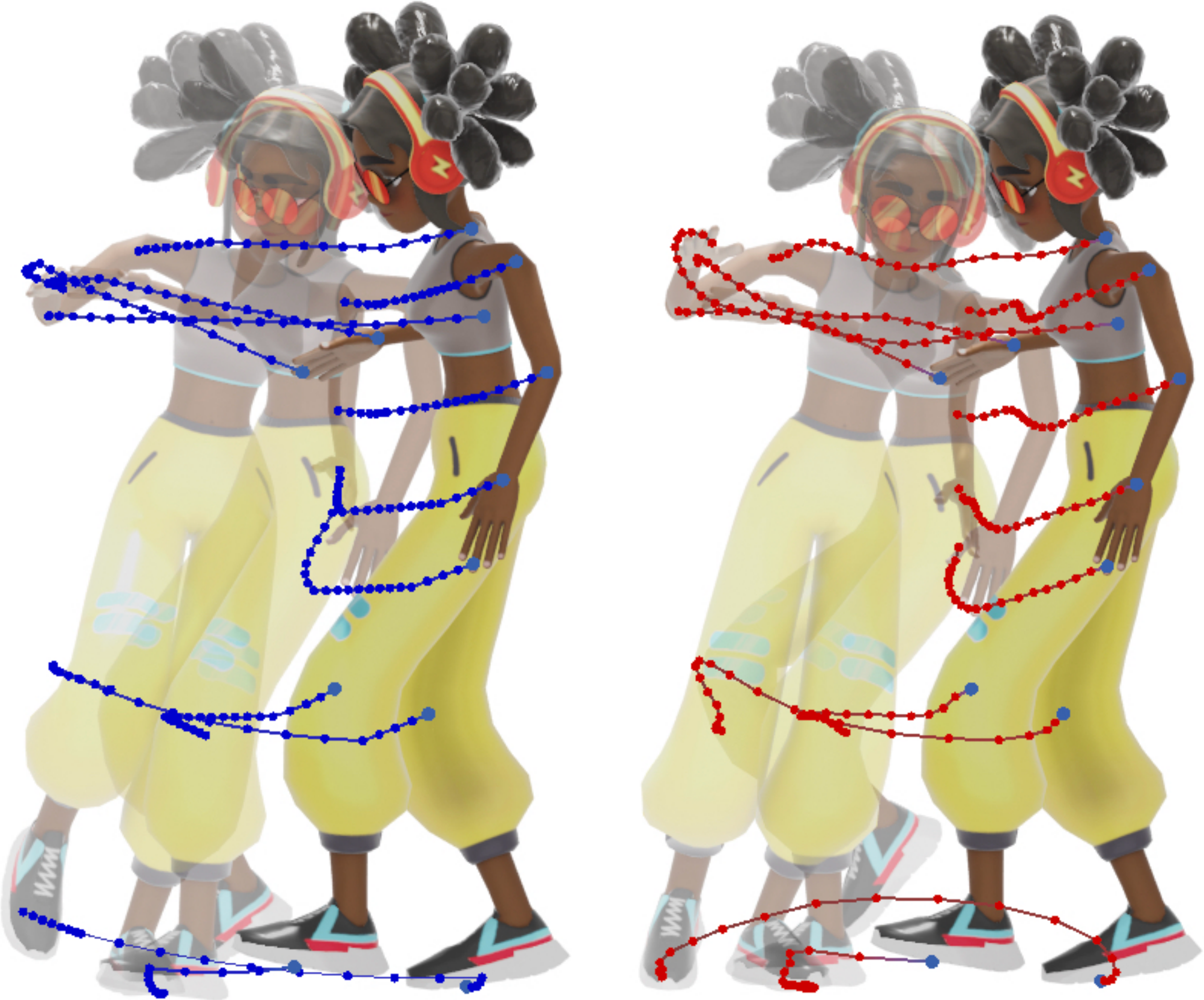}
    \caption{Lagrangian Motion Fields of the recovered motion (represented in blue) exhibit unnatural stiffness and a deficiency in detail. To address this limitation, a lightweight refinement module is applied to improve both realism and diversity. The refined motion (illustrated in red) retains finer details and exhibits more natural transitions between frames.}
    \label{fig:motion_supermotion_flow}
\end{figure}

After generating supermotions, we reconstruct the complete motion sequence for any given time within each supermotion using~\refeq{sm2m}. In contrast to previous methods~\cite{lodge} that produce sparse keyframes based on prior knowledge and manually designed rules, our approach reconstructs full motion sequences directly from supermotions without relying on predefined constraints, thereby enhancing efficiency, conciseness, and generalizability.

% \input{tab/tab_rec_comp}
% \revise{
% However, as shown in \reftab{vqvae_comparison}, while learning-based motion compression methods like VQ-VAE and its K-means variant achieve better reconstruction quality, our Lagrangian Motion Fields approach trades some reconstruction accuracy for physical interpretability and direct motion control capabilities.
% }
However, as shown in~\reffig{motion_supermotion_flow}, the recovered motion sequences may exhibit unnatural stiffness and lack of detail due to approximation errors, leading to imprecise motion expressions and reduced diversity.
To address these issues, we implement a lightweight diffusion-based refinement module to enhance high-frequency motion details and increase diversity.
Specifically, we train a conditional diffusion model $G$ that takes the recovered motion sequence $\hat{\mathbf{X}}$ as conditions and predicts the clean motion sequence $\mathbf{X}$. 
The refine loss $\mathcal{L}_{refine}$ is defined as follows:
\begin{equation} \label{diffusion_forward}
    q(\mathbf{X}_\tau|\mathbf{X}_{\tau-1}) = \mathcal{N}(\mathbf{X}_\tau;\sqrt{1- \beta_\tau}\mathbf{X}_{\tau-1}, \beta_\tau\mathbf{I}).
\end{equation}
\begin{equation} \label{eq:refinment}
    \mathcal{L}_{refine} = \mathbb{E}_{\mathbf{X}, \tau}{\lVert \mathbf{X} - G([\mathbf{X}_\tau, \hat{\mathbf{X}}],\tau) \rVert_2^2}
\end{equation}
where $\beta_\tau$ denotes the pre-defined noise variance schedule, $[\cdot]$ denotes concatenation. To avoid any confusion, we use $\tau$ to denotes the diffusion denoising timestep, distinguishing it from $t$, which represents the frame index.

\subsection{Adaptation to Downstream Tasks} \label{sec:adaptation}
Our motion representations can be applied to a variety of motion-related downstream tasks.
To demonstrate the effectiveness and generalizability of our approach, we adapt it to two multi-modal conditional motion generation tasks: music-to-dance and text-to-motion.

\textbf{Music-to-Dance.} We begin by analyzing all dance motions through clustering velocities at each timestep using K-means, which generates supermotions corresponding to each music piece. 
{We select a generation window size $L$ and a sampling stride $S$ for the supermotions. 
The corresponding music signal window size $L^{\prime}$ is set to be the average length of the recovering motion length from $L$ supermotions follwing~\refeq{sm2m}.

Given a lengthy music control signal with the length of $L_m$, we first pad it to ensure its length becomes $n$ multiple of $L^{\prime}$. We then apply the long-form sampling method proposed by Tseng et. al.~\cite{edge} to generate $n$ supermotion sequences with length of $L$. 
After recovery following~\refeq{sm2m} and refinement in~\refsec{refinement_module}, we obtain a total motion sequence of length $L^{\prime}_m$, where $L^{\prime}_m$ may vary. 
Empirically, $L$ and $L^{\prime}$ are chosen such that the duration $L$ of supermotions statistically approximates the length $L^{\prime}$ of motions. 
The predicted motion sequence is then clipped or interpolated to match the exact length $L_m$ of the music.

\textbf{Text-to-Motion.} The preprocess for text-to-motion follows the same steps as for music-to-dance. The motion sequence is converted into supermotion sequences, with the text condition length padded to match the longest text prompt in the training dataset. 
All motion sequences are converted into supermotion sequences, and $L^{\prime}$ is selected so that in over 90\% of cases, the supermotion window $L$ covers the corresponding motion window $L^{\prime}$. 
For constructing the refinement dataset, if the duration of the recovered motion exceeds the motion length, we pad the motion using the final frame motion, corresponding to stationary motion.

In both downstream tasks, the refinement window size is set as $L_r$, which is significantly smaller than $L^{\prime}$. 
During inference, the recovered motion is divided into non-overlapping segments, which are refined parallelly. 
The refined segments are then concatenated to form the final result.}

\section{Experiment} \label{sec:experiment}
In this section, we conduct rigorous evaluations of our approach for long-term music-to-dance and long-term text-to-motion, incorporating both quantitative in~\refsec{music_to_dance} and~\refsec{text_to_motion}, as well as qualitative analyses in~\refsec{qualitative_analyses}.
To get a more comprehensive insight into our main contributions, we conduct ablation studies in~\refsec{ablation_study} to validate the effectiveness of the proposed pipeline.

\subsection{Long-term Music-to-Dance} \label{sec:music_to_dance}
The task of music-to-dance generation involves generating dance motions synchronized with the input music.
In this subsection, we implement our method based on the two SOTA diffusion-based music-to-dance methods, EDGE~\cite{edge} and Lodge~\cite{lodge}.
We use the same network architecture and sequence window length as these methods for a fair comparison. 
During training, the motion representation is substituted with supermotion, and the origin loss terms are replaced with the newly proposed loss terms specifically designed for supermotion.

\textbf{Datasets.}
We conduct the long-term music-to-dance evaluation using the FineDance dataset~\cite{finedance}, which comprises 7.7 hours of 30 FPS music and dance motion pairs, covering 16 dance genres, totaling 831,600 frames.
The average duration of each dance segment is 152.3 seconds.
In contrast to the commonly used AIST++ dataset~\cite{fact}, which has an average duration of only 13.3 seconds, FineDance better aligns with our long-term setting and poses a greater challenge.
During preprocessing, we employ mini-batch K-Means with 1,000 clusters.

\subsubsection{Implementation details}
For the supermotion generation module, the length of the input music condition is set to 1100 and the output length of supermotion is set to 150.
For loss weight hyperparameters, we set $\lambda_{recon} = 0.636$, $\lambda_{joint} = 0.646$, $\lambda_{vel} = 0.0$, $\lambda_{contact} = 10.942$, and $\lambda_{coherent} = 2.964$ when training supermotion based on EDGE. $\lambda_{recon} = 0.636$, $\lambda_{joint} = 0.636$, $\lambda_{vel} = 2.964$, $\lambda_{contact} = 10.942$, and $\lambda_{coherent} = 2.964$ while based on Lodge.
We train the model using the Adan optimizer~\cite{adan} with a learning rate of 4e-4. 
The batch size is 384. 
The supermotion model is trained for 8,000 epochs and the refinement module is trained for 4,000 epochs. 
The training process takes 48 hours on 6 NVIDIA L40 GPUs for the supermotion module and 22 hours for the refinement module.

\subsubsection{Quantitative Evaluation}

We compare our method with state-of-the-art works, including FACT~\cite{fact}, MNET~\cite{mnet}, Bailando~\cite{bailando}, EDGE~\cite{edge}, and Lodge~\cite{lodge}.
We conduct comprehensive evaluations based on three aspects: motion quality, diversity, and computational efficiency.

\begin{table*}[ht] 
    \centering
    \caption{Comparison with SOTAs on the FineDance dataset. \textbf{Bold} numbers indicate the best performance among all methods. The underlined numbers show improvements achieved by our method compared to baseline methods.}
    \label{tab:m2d_comparison}
    \footnotesize
    \begin{tabular}{l cccc cc c}
        \toprule[1.5pt]
        \multirow{2}*{Method} & \multicolumn{4}{c}{Motion Quality} & \multicolumn{2}{c}{Motion Diversity} & \multicolumn{1}{c}{Efficiency} \\ 
        \cmidrule(lr){2-5} \cmidrule(lr){6-7} \cmidrule(lr){8-8}
        & BAS$\uparrow$ & FSR $\downarrow$ & $\mathrm{FID}_k\downarrow$ & $\mathrm{FID}_g\downarrow$ & $\mathrm{Div}_k\uparrow$ & $\mathrm{Div}_g\uparrow$ & Runtime$\downarrow$ \\
        \midrule
        Ground Truth &0.2120 & 6.22$\%$ & / & / & 9.73 & 7.44  & / \\
        \midrule
        FACT~\cite{fact} & 0.1831 & 28.44$\%$ & 113.38 & 97.05 & 3.36 & 6.37 & 35.88s \\
        MNET~\cite{mnet} & 0.1864 & 39.36$\%$ & 104.71 & 90.31 & 3.12 & 6.14 & 38.91s \\
        Bailando~\cite{bailando} & 0.2029 & 18.76$\%$ & 82.81 & 28.17 & 7.74 & 6.25 & 5.46s \\
        EDGE~\cite{edge} & 0.2116 & 20.04$\%$ & 94.34 & 50.38 & 8.13 & \textbf{6.45} & 8.59s \\
        Lodge (DDIM)~\cite{lodge} & 0.2269 & 2.76$\%$ & \textbf{50.00} & 35.52 & 5.67 & 4.96 & 4.57s \\
        \midrule
        \revise{Long-term Bailando~\cite{bailando} + K-Means} & 0.2134 & 18.20$\%$ & 72.52 & 26.70 & 7.12 & 6.01 & 5.40s \\
        \midrule
        EDGE + Ours (3D)  & \underline{\textbf{0.2350}} & 24.91$\%$ & \underline{50.27} & \underline{31.63} & 5.63 & 5.27 & \underline{\textbf{2.48}s} \\
        EDGE + Ours (6D)  & \underline{0.2288} & $\underline{12.63 \%}$ & \underline{55.49} & \underline{24.93} & 5.46 & 5.82 & \underline{2.51s} \\
        Lodge (DDIM) + Ours (3D) & \underline{0.2348} & 3.94$\%$ & 52.98 & \underline{29.17} & 5.41 & \underline{5.62} & \underline{2.53s} \\
        Lodge (DDIM) + Ours (6D) & \underline{0.2283} & \underline{\textbf{1.45$\%$}} & 62.16 & \underline{\textbf{23.39}} & 4.95 & \underline{6.18} & \underline{2.57s} \\
        \bottomrule[1.5pt]
    \end{tabular}
\end{table*}

Following previous work, we assess the motion quality from four different perspectives: Beat Alignment Score (BAS)~\cite{fact, mnet, bailando, edge, lodge}, Foot Skating Ratio (FSR)~\cite{guided-motion}, and Fréchet Inception Distance (FID) of kinetic ($\mathrm{FID}_k$) and geometric ($\mathrm{FID}_g$) following~\cite{lodge}. BAS measures the synchronization between the generated dance and the given music. FSR evaluates the proportion of foot sliding in the generated motion, reflecting physical realism. $\mathrm{FID}_k$ assesses the physical realism of the motion through speed and acceleration features. $\mathrm{FID}_g$ measures the overall choreography quality based on predefined geometric motion templates.
For assessing the diversity, we measure the average feature distance of kinetic ($\mathrm{Div}_k$) and geometric ($\mathrm{Div}_g$) features extracted by fairmotion~\cite{gopinath2020fairmotion} following the previous works~\cite{bailando}.
Efficiency is evaluated by measuring the average runtime required to generate a dance sequence of 1024 frames.

To further demonstrate the robustness of the proposed method, we provide quantitative results for two most commonly used coordinate systems: Cartesian coordinates (denoted as ``3D") and 6-DOF rotation coordinates (denoted as ``6D"). These results offer a more thorough comparison, which we will discuss more in~\refsec{ablation_study}.
The quantitative results, detailed in~\reftab{m2d_comparison}, demonstrate significant improvements achieved by our method. 
Notably, our approach shows substantial advancements in $\mathrm{FID}_g$ scores, with ``Lodge (DDIM) + Ours (6D)" achieving the highest score of 23.39, reflecting superior motion quality. 
Additionally, ``Lodge (DDIM) + Ours (6D)" records the lowest Foot Skating Ratio of 1.45, underscoring the model's capability to produce natural and physically plausible motions.
Furthermore, ``EDGE + Ours (3D)" achieves the highest Beat Alignment Score (BAS) of 0.235, demonstrating superior synchronization with music. 

Our method demonstrates improvements in motion diversity, particularly in the geometric feature space, as we got higher $\mathrm{Div}_g$ scores across multiple configurations compared to baseline methods.
Most importantly, our method drastically reduces the runtime due to the superior performance of our proposed supermotion representation, achieving a maximum acceleration of 71.13\% based on EDGE and 44.64\% based on Lodge.

\revise{
\subsubsection{Comparison with Learned Compact Representations}
Bailando~\cite{bailando} is a VQ-VAE-based motion generation method that encodes motion into discrete latent codes for reconstruction. This approach can be extended for long-term representation learning by applying K-Means clustering to latent codes. To evaluate this, we implement a variant, ``Long-term Bailando+K-Means'', as shown in~\reftab{m2d_comparison}, where we cluster the re-trained encoder outputs of Bailando for codebook initialization. This follows the VQ-VAE training practices in~\cite{lancucki2020robust}, incorporating proper codebook initialization and training stability optimizations.

However, directly clustering these latent codes fails to account for motion dynamics, velocity continuity, and biomechanical constraints. As shown in~\reftab{m2d_comparison}, our method outperforms ``Long-term Bailando+K-Means'' across key metrics, demonstrating superior long-term motion generation. While VQ-VAE effectively reconstructs training samples, our supermotion representation better preserves motion coherence by structuring motion based on velocity patterns rather than discrete latent codes.
}

\subsection{Long-term Text-to-Motion} \label{sec:text_to_motion}
Text-to-motion aims at generating reasonable motion sequences from text descriptions. 
% Normal T2M
We utilize MDM~\cite{mdm} as the backbone model and conduct comparison experiments with the original MDM, JL2P~\cite{JL2P}, Text2Gesture~\cite{text2gestures}, and T2M~\cite{T2M}.
% Long-term T2M
To more effectively demonstrate the superiority of our representation in long-term text-to-motion modeling, we further employ MDM and T2M-GPT~\cite{t2m-gpt} as backbone models and conduct additional comparison experiments on our synthetic multi-prompts text-to-motion dataset.

\subsubsection{Long-term Text-to-Motion Dataset HumanML3D-MP}
We conduct the ordinary text-to-motion evaluation using HumanML3D dataset~\cite{T2M} as previous work \cite{mdm, t2m-gpt}. 
HumanML3D comprises 14,616 motions with 44,970 text scripts in total, sourced from the AMASS~\cite{amass} and HumanAct12~\cite{humanact12} datasets. 
The motion in this dataset covers a wide range of human activities, such as exercising, acrobatics, and dancing. 
However, the average duration of each motion segment is 7.1 seconds, and the longest segment is only 10 seconds, making it unsuitable for evaluation for long-term text-to-motion generation.

To more effectively assess the performance of our method in long-term motion generation, we constructed a multi-prompt text-to-motion dataset based on HumanML3D, dubbed HumanML3D-MP.
HumanML3D-MP consists of 20,000 motions paired with 60,000 text scripts. Each motion segment has an average duration of 74.8 seconds, making it a premier benchmark for long-term motion generation.

Specifically, we first select motion sequences from HumanML3D with lengths ranging between 40 and 200, removing all short sequences. Subsequently, we stitch the motion sequences and corresponding text. The stitch strategy for the motion sequences is as follows. The original motion representation $\omega=(R,T)$ in HumanML3D uses the 6-DOF rotation representation, denoted as $R \in \mathbb{R}^{L \times (22 \times 6)}$, for every joint, and a single root translation is denoted by $T \in \mathbb{R}^{L \times 3}$. For stitching $\omega_1$ and $\omega_2$, denoted as $\omega = \mathbf{Stitch}(\omega_1, \omega_2)$, we first compute the relative displacement between frames, $\Delta T_1 = T_1[1:] - T_1[:-1]$ and $\Delta T_2 = T_2[1:] - T_2[:-1]$, and then select the first frame of $T_1$ as the initial root translation, i.e., $T[0] = T_1[0]$. The subsequent root translations are computed based on $\Delta T_1$ and $\Delta T_2$, as follows:
\begin{equation}
T[i+1] =
\begin{cases}
T[i] + \Delta T_1[i], i \in [0, L_1-1] \\
T[i] + \Delta T_2[i-L_1], i \in [L_1, L_1+L_2-1].
\end{cases}
\end{equation}
For the rotation representation, we need to make the transition between each motion smooth, so we use a fade-in and fade-out interpolation within the transition range $M(=20)$. The strategy involves selecting the last $M$ frames of the preceding motion and the first $M$ frames of the succeeding motion, then applying the fade-in and fade-out method for interpolation. Specifically, let $R_1$ represent the rotation data for the preceding motion and $R_2$ for the succeeding motion. We define the fade-in function and fade-out function, and the interpolated rotation for the transition frames is then computed as follows:
\begin{equation}
    R[i] = \begin{cases} 
        R_1[I] &, i \in I_1\\
        f_{\text{out}}(j) R_1[L_1 - M + j] + f_{\text{in}}(j) R_2[j] &, i \in I_2\\
        R_2[i - L_1] &, i \in I_3
    \end{cases}
\end{equation}
, where
\begin{equation}
    \begin{aligned}
    & I_1 = [0, L_1 - M - 1], \\
    & I_2 = [L_1 - M, L_1 - 1], \\
    & I_3 = [L_1, L_1 + L_2 - M - 1], \\
    & f_{\text{in}}(j) = \frac{j}{M}, \quad f_{\text{out}}(j) = \frac{M-j}{M}, \\
    & j = i - (L_1 - M). \\
    \end{aligned}
\end{equation}

For building a sample, we randomly select 10 motions $\omega_1, \omega_2, ..., \omega_{10}$ and concatenate these motions sequentially as below:
\begin{equation}
    \begin{aligned}
        & \omega^1 = \omega_1, \omega^{k+1} = \mathbf{Stitch}(\omega^k, \omega_{k+1}), \\
        & \omega = \omega^{10},
    \end{aligned}
\end{equation}
and get $\omega$ as our final stitched motion. For text prompts, we assume that all motions are performed by the same person. Therefore, we unify the subject of the first text prompt as ``The person''.
We then use part-of-speech tagging to locate the position of the subject in subsequent text prompts and replace all subsequent subjects with ``And then this person''.
Finally, we concatenate these sentences to obtain our final text prompt.

In the original dataset, each motion is associated with multiple text prompts. 
We randomly select one text description for each short motion and then concatenate 10 descriptions to create our text prompt for stitched motion. 
In HumanML3D-MP, we build 20,000 motions and construct three text prompts for each motion.

\subsubsection{Implementation Details}
\textbf{MDM Variant.}
For the supermotion generation module, the output length of the supermotion is set to 40 for HumanML3D and 400 for HumanML3D-MP. Supermotion sequences are generated in a single pass for all text conditions. The loss weights are configured as follows: $\lambda_{recon} = 1.0$, $\lambda_{joint} = 0.0$, $\lambda_{vel} = 0.0$, $\lambda_{contact} = 0.0$, and $\lambda_{coherent} = 0.2$ to align with MDM's settings. The model is trained for 4,000 epochs with a batch size of 192 using the Adam optimizer, with a learning rate of 1e-4. The training process takes approximately 11 hours on 6 NVIDIA L40 GPUs for HumanML3D and 48 hours for HumanML3D-MP.

\textbf{T2M-GPT Variant.}
Instead of using the neural-based VQ-VAE to obtain the motion latent representation as in the original T2M-GPT~\cite{t2m-gpt}, we utilize the proposed supermotion representation as the intermediate result. For a fair comparison, our supermotion sequence uses a clustering model derived from HumanML3D and employs the same transformer architecture, which includes 18 transformer layers with a dimension of 1,024 and 16 heads. The model is trained for 300,000 iterations with a batch size of 192 using the AdamW optimizer, starting with a learning rate of 1e-4 for the first 150,000 iterations, which is then decayed to 5e-6 for the remaining 150,000 iterations. The entire training process takes 14 hours for HumanML3D-MP on 6 NVIDIA L40 GPUs.

\begin{table*}[t] 
    \centering    
    \caption{Quantitative results on the HumanML3D test set. All methods use the real motion length from the ground truth. `$\uparrow$' indicates the higher the better. `$\downarrow$' indicates the lower the better.`$\rightarrow$' indicates that results are better when the metric is closer to the real distribution. The evaluations were performed 20 times, with $\pm$ indicating the 95\% confidence interval. \textbf{Bold} indicates the best result, and underline indicates it surpasses its baseline.}
    \footnotesize
    \begin{tabular}{l ccccc}
        \toprule[1.5pt]
        \multirow{2}{*}{Method} & \multicolumn{3}{c}{Motion Quality} & \multirow{2}{*}{Diversity$\rightarrow$} & \multirow{2}{*}{Runtime$\downarrow$} \\
        \cmidrule(lr){2-4}
        & R Precision (top 3)$\uparrow$ & FID$\downarrow$ & Multimodal Dist$\downarrow$ \\
        \midrule
        Ground Truth & $0.797^{\pm.002}$ & $0.002^{\pm.000}$ & $2.974^{\pm.008}$ & $9.503^{\pm.065}$ & -\\ 
        \midrule
        JL2P &  $0.486^{\pm.002}$ & $11.02^{\pm.046}$ & $5.296^{\pm.008}$ & $7.676^{\pm.058}$ & - \\
        Text2Gesture &  $0.345^{\pm.002}$ & $7.664^{\pm.030}$ & $6.030^{\pm.008}$ & $6.409^{\pm.071}$ & - \\
        T2M & ${0.740^{\pm.003}}$ & $1.067^{\pm.002}$ & $\textbf{3.340}^{\pm.008}$ & $9.188^{\pm.002}$ & - \\
        MDM &  ${0.611^{\pm.007}}$ & $\textbf{0.544}^{\pm.044}$ & ${5.566^{\pm.027}}$ & ${\textbf{9.559}^{\pm.086}}$ & 32.30s \\
        \midrule
        MDM + Ours (3D) &  \underline{$0.737^{\pm0.004}$} & ${1.736^{\pm0.029}}$ & \underline{${3.4468^{\pm0.030}}$}& ${9.578^{\pm0.057}}$  & \underline{7.8s} \\
        MDM + Ours (6D) &  \underline{$\textbf{0.757}^{\pm0.007}$} & ${1.138^{\pm0.027}}$ & \underline{${3.5620^{\pm0.006}}$}& $9.425^{\pm0.081}$  & $\underline{\textbf{7.1s}}$ \\
        \bottomrule[1.5pt]
    \end{tabular}
    \label{tab:t2m_comparison}
\end{table*}
\begin{table*}[t]
    \centering
    \renewcommand{\arraystretch}{1.2}  % 增加行距
    \caption{Comparison with the baseline methods on the HumanML3D-MP test set. For each metric, we repeat the evaluation 20 times and report the average with a 95\% confidence interval. The \textbf{bold} number is the best result in the same group.}
    \footnotesize
    \begin{tabular}{l cccc c c c}
        \toprule[1.5pt]
        \multirow{2}{*}{Method}  & \multicolumn{3}{c}{R-Precision $\uparrow$} & \multirow{2}{*}{FID$\downarrow$} & \multirow{2}{*}{Multimodal Dist$\downarrow$} & \multirow{2}{*}{Diversity$\rightarrow$} & \multirow{2}{*}{Runtime$\downarrow$} \\
        \cline{2-4}
        ~ & Top-1 & Top-2 & Top-3 \\
        \midrule
        Ground Truth & $0.321^{\pm 0.005}$ & $0.468^{\pm 0.005}$ & $0.565^{\pm 0.006}$ & $0.002^{\pm 0.001}$ & $4.204^{\pm 0.024}$ & $6.618^{\pm 0.011}$ & - \\
        \midrule
        MDM & $0.155^{\pm 0.002}$ & $0.260^{\pm 0.003}$ & $0.340^{\pm 0.002}$ & $2.315^{\pm 0.110}$ & $5.033^{\pm 0.018}$ & $4.776^{\pm 0.047}$ & 218.34s \\
        MDM + Ours (3D) & $0.201^{\pm 0.003}$ & $0.338^{\pm 0.004}$ & $0.442^{\pm 0.005}$ & $1.668^{\pm 0.027}$ & $\textbf{4.610}^{\pm 0.019}$ & $\textbf{5.842}^{\pm 0.039}$ & 11.83s\\
        MDM + Ours (6D) & $\textbf{0.213}^{\pm 0.005}$ & $\textbf{0.351}^{\pm 0.006}$ & $\textbf{0.455}^{\pm 0.007}$ & $\textbf{1.626}^{\pm 0.031}$ & $4.720^{\pm 0.023}$ & $5.537^{\pm 0.048}$ & 11.66s \\
        \midrule
        T2M-GPT & $0.170^{\pm 0.002}$ & $0.286^{\pm 0.003}$ & $0.374^{\pm 0.002}$ & $2.083^{\pm 0.030}$ & $4.800^{\pm 0.027}$ & $5.415^{\pm 0.056}$ & 2.30s\\
        T2M-GPT + Ours (3D) & $0.204^{\pm 0.004}$ & $0.343^{\pm 0.004}$ & $0.453^{\pm 0.005}$ & $1.618^{\pm 0.027}$ & $\textbf{4.650}^{\pm 0.024}$ & $\textbf{5.774}^{\pm 0.061}$ & 0.54s \\
        T2M-GPT + Ours (6D) & $\textbf{0.217}^{\pm 0.004}$ & $\textbf{0.359}^{\pm 0.006}$ & $\textbf{0.470}^{\pm 0.005}$ & $\textbf{1.586}^{\pm 0.021}$ & $4.550^{\pm 0.020}$ & $5.632^{\pm 0.045}$ & 0.53s \\
        \bottomrule[1.5pt]
    \end{tabular}
    \label{tab:t2m_mp}
\end{table*}

\subsubsection{Quantitative Evaluation}
We evaluate the generated motions based on quality, accuracy, diversity, and efficiency, adhering to the evaluation protocol outlined in MDM~\cite{mdm}. 
For measuring motion quality, we employ FID~\cite{humanact12}, R-Precision, and Multimodal Distance~\cite{T2M}. 
R-Precision measures the accuracy of motion-to-text retrieval by ranking the Euclidean distances between motion and text embeddings, with results reported for Top-1, Top-2, and Top-3 accuracy. 
FID evaluates the distribution distance between generated and real motions by comparing the extracted motion features. 
Multimodal Distance calculates the average Euclidean distance between text features and their corresponding generated motion features, assessing the alignment between text and motion.
To assess motion diversity, we assess the variation within a set of generated motions by computing the average Euclidean distance between randomly sampled pairs of motion features, following ~\cite{humanact12}. The runtime metric measures the time required to sample one motion instance.
For a more comprehensive analysis, we conduct comparisons on both short-term and long-term text-to-motion benchmarks, as presented in~\reftab{t2m_comparison} and~\reftab{t2m_mp}, respectively.

As illustrated in~\reftab{t2m_comparison}, the incorporation of our representation significantly enhances MDM's performance in terms of quality, accuracy, and efficiency. 
Specifically, our method increases R-Precision (Top-3) from 0.611 to 0.737 (3D) and 0.757 (6D), representing improvements of 20.6\% and 23.9\%, respectively. 
Additionally, it improves the Multimodal Distance by up to 38.1\%, demonstrating our representation's effectiveness in enhancing the model's ability to understand the correlation between text and motion.

Notably, our method demonstrates a more pronounced advantage in long-term text-to-motion tasks. As detailed in ~\reftab{t2m_mp}, our methods outperform its baseline models in both R-Precision and Multimodal Distance on the HumanML3D-MP dataset. Specifically, FID decreases by 28\% compared to MDM and by 24\% compared to T2M-GPT, indicating a significant improvement in motion quality.
In addition, our method shows improvements in motion diversity, increasing by 22\% compared to MDM and by 7\% compared to T2M-GPT.

The compact nature of our supermotion representation also contributes to a 75\% reduction in runtime compared to the original MDM on HumanML3D. Moreover, while T2M-GPT incurs additional training costs for VQVAE, our supermotion compression model imposes negligible costs, further highlighting the efficiency of our approach.
These comprehensive improvements across quality, accuracy, and diversity metrics confirm the effectiveness of our approach, further establishing its superiority over the baseline methods in generating realistic and diverse long-term motions.
\begin{figure*}[t]
   \centering
   \includegraphics[width=\textwidth]{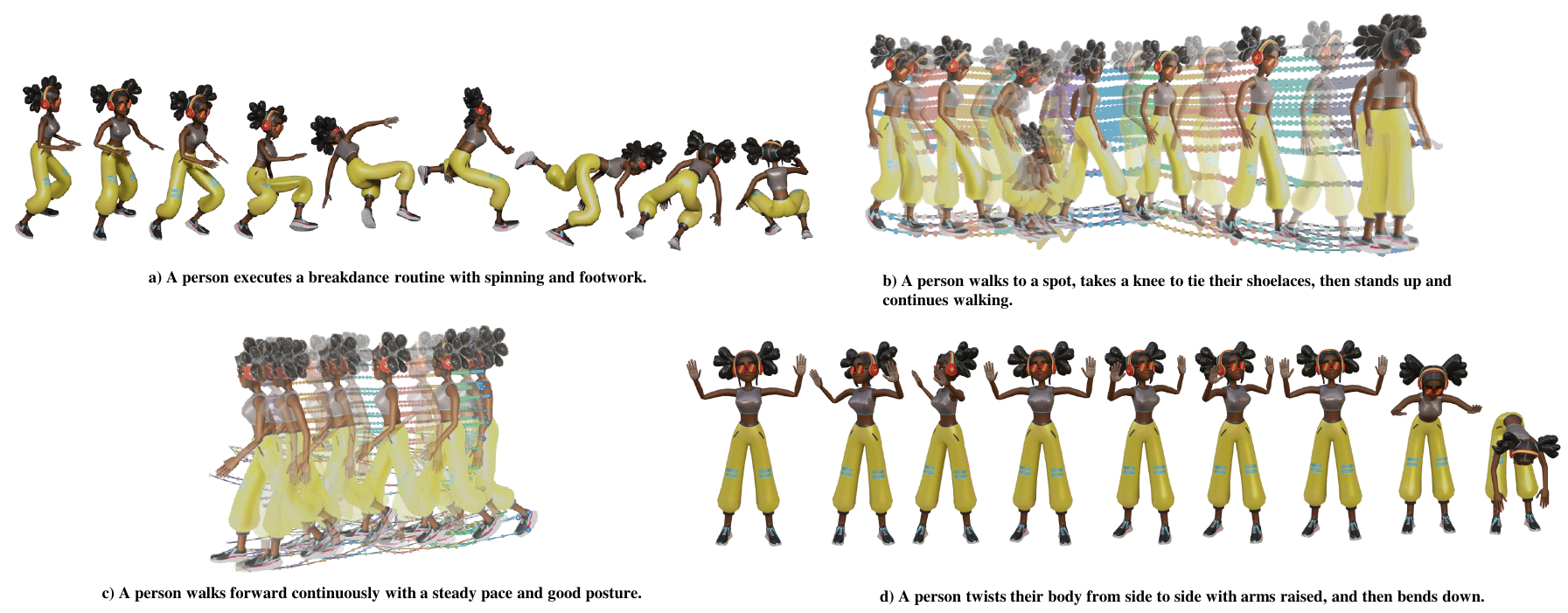}
   \caption{
   \revise{
Qualitative visualization of our method across different motion types: (a) Dance movements; (b) Multi-phase sequential actions; (c) Cyclic locomotion;
(d) Large-amplitude steady motions.
   }
   }
   \label{fig:motion_type}
\end{figure*}
\begin{figure*}[t]
   \centering
   \includegraphics[width=\linewidth]{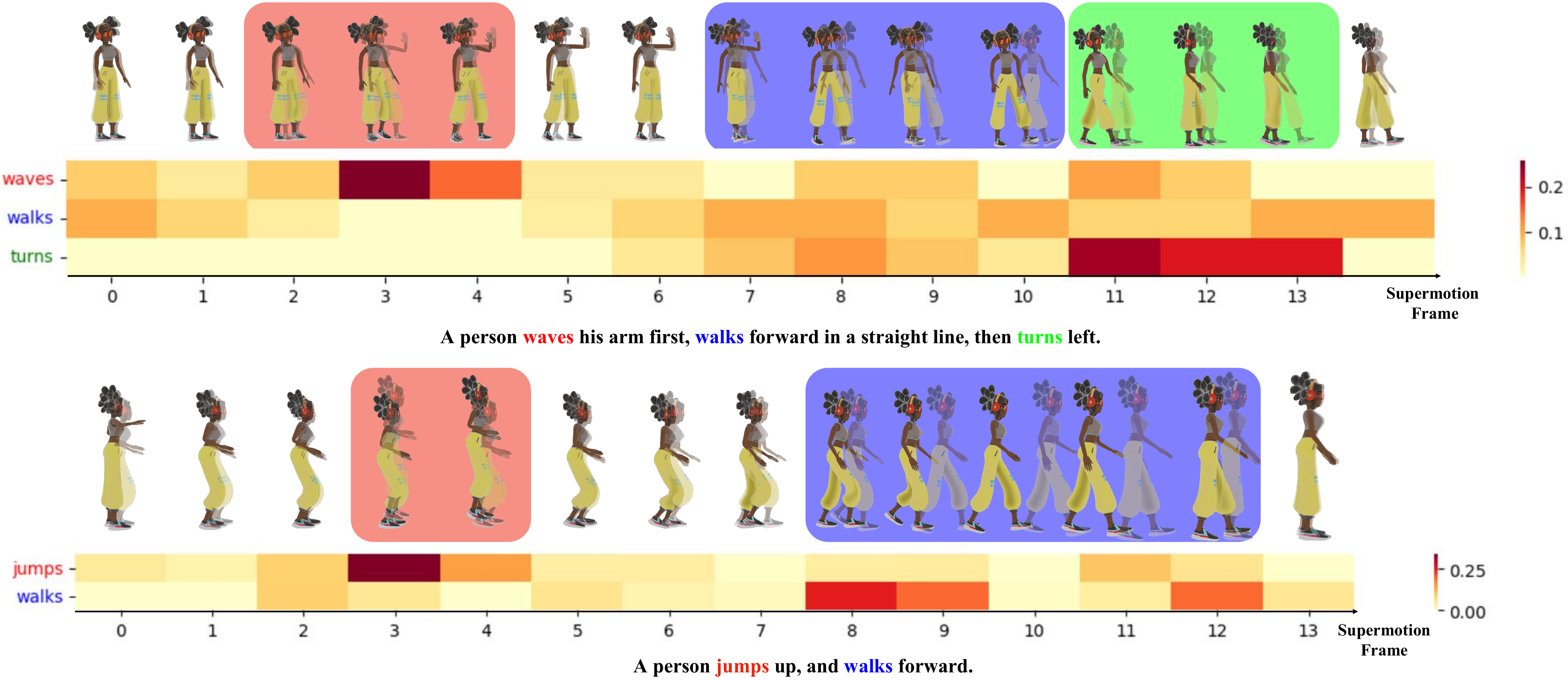}
   \caption{\revise{Text-supermotion alignment visualization. Each example presents the motion sequence (top) and the corresponding attention heatmap (bottom), illustrating the attention weights between text tokens and temporal segments. Darker colors indicate stronger attention weights.}}
   \label{fig:attention}
\end{figure*}
\begin{figure*}[t]
    \centering
    \includegraphics[width=\textwidth]{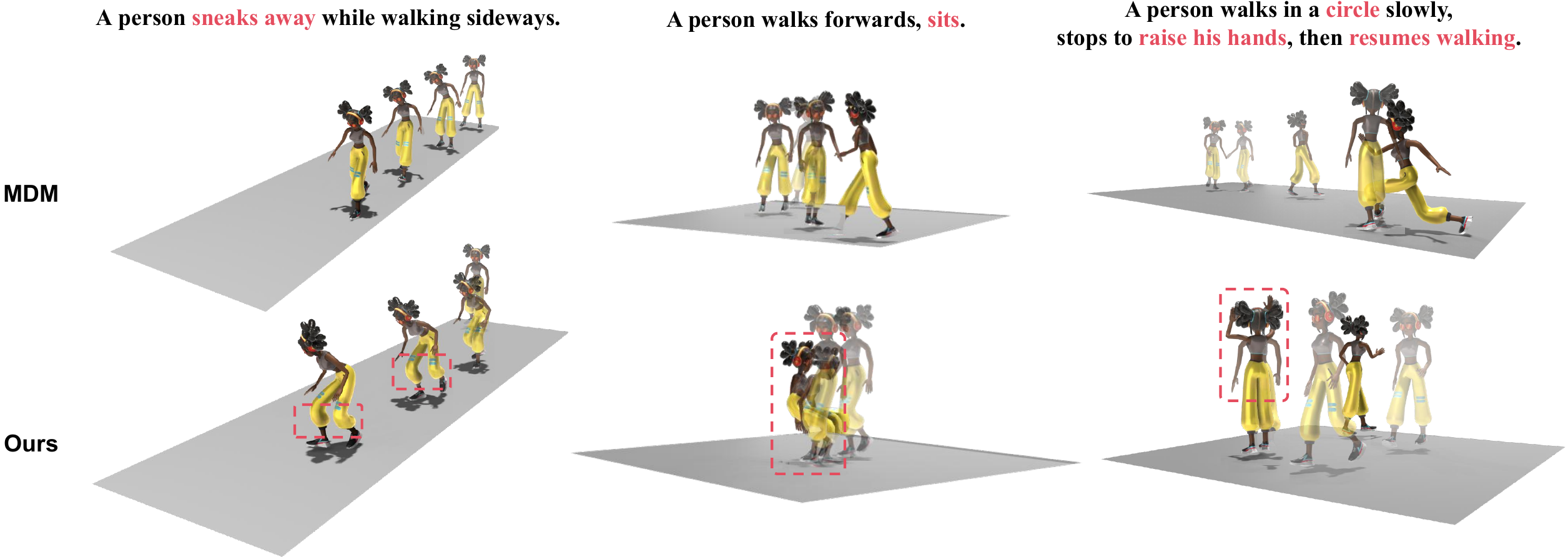}
    \caption{Qualitative comparison of text-to-motion results. Video demonstrations are available in the supplementary materials.}
    % \vspace{-4mm}
    \label{fig:t2m-vis}
\end{figure*}
\begin{figure}[t]
    \centering
    \includegraphics[width=\linewidth]{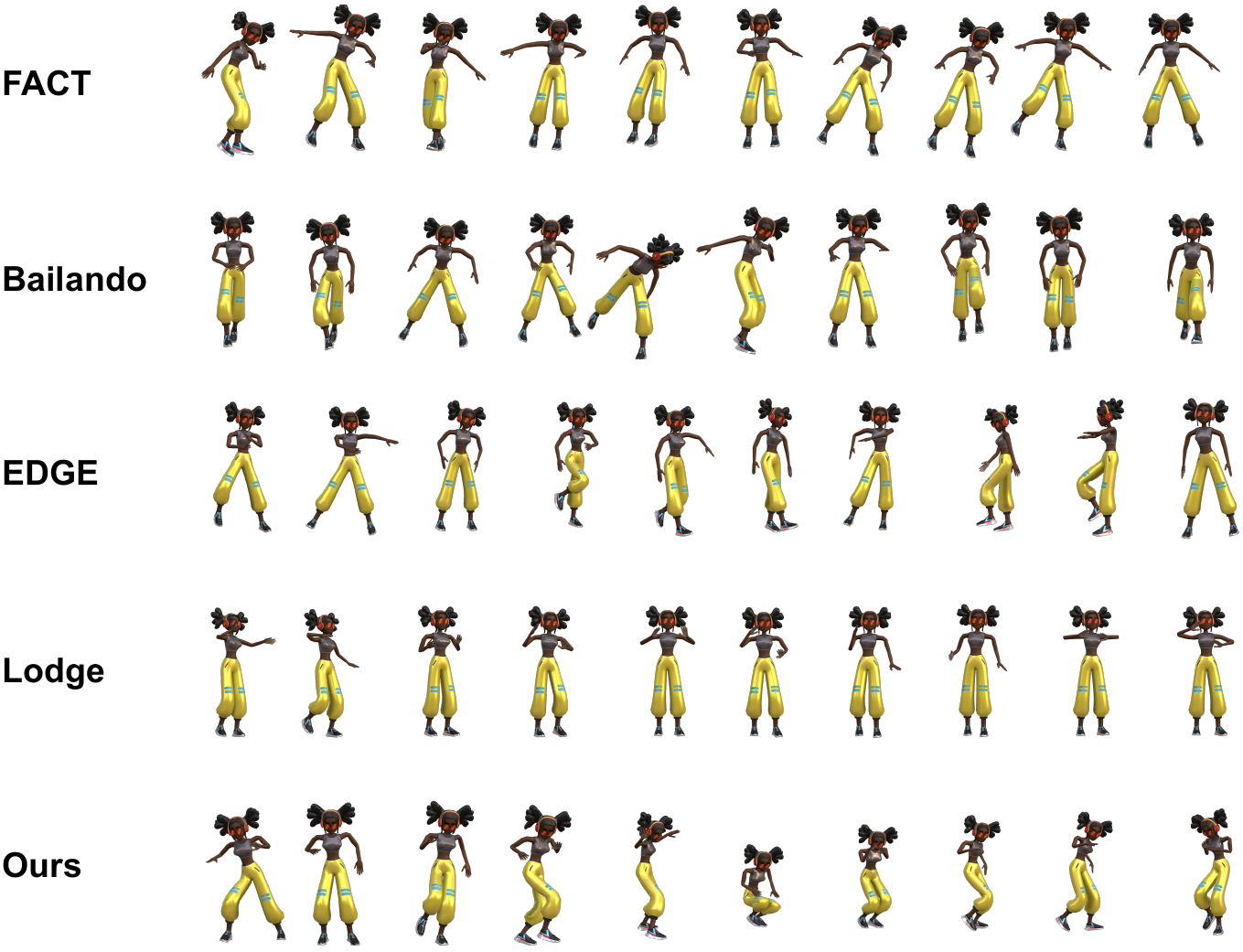}
    % \vspace{-4mm}
    \caption{Qualitative comparison of music-to-dance results. Video demonstrations are available in the supplementary materials.}
    % \vspace{-4mm}
    \label{fig:m2d-vis}
\end{figure}
\begin{figure}[t]
    \centering
    \begin{tikzpicture}
        \begin{axis}[
            title={User Study},
            symbolic x coords={Lodge, Bailando, EDGE, FACT, T2M-GPT, MDM},
            xtick=data,
            ylabel=Win Rate,
            ymin=0, ymax=1,
            bar width=10pt,
            enlarge x limits=0.3,
            yticklabel={\pgfmathparse{\tick*100}\pgfmathprintnumber{\pgfmathresult}\%},
            nodes near coords,
            ybar,
            width=1\linewidth,
            height=0.7\linewidth,
            x tick label style={rotate=45,anchor=east}, % 旋转x轴标签
            xlabel style={yshift=-10pt}, % 调整x轴标题位置
        ]
        \addplot coordinates {(Lodge,0.727) (Bailando, 0.864) (EDGE, 0.789) (FACT, 0.674) (T2M-GPT, 0.802) (MDM, 0.818)};
        \end{axis}
    \end{tikzpicture}
    \caption{Pairwise user preference study results. ``Win Rate'' represents the ratio at which each method is preferred in the video pairs.}
    \label{fig:user_study}
\end{figure}
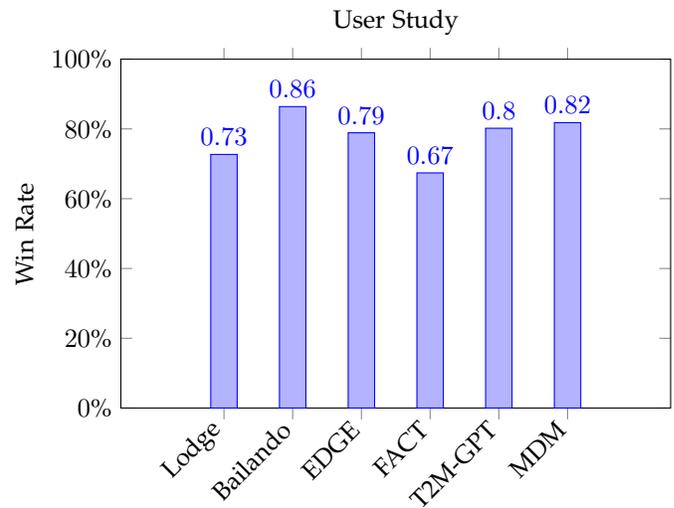
\subsection{Qualitative Evaluation}  \label{sec:qualitative_analyses}
\revise{
\subsubsection{Analysis of Motion Types} \label{sec:motion_type}  

% As shown in~\reffig{motion_type}, our approach exhibits varying performance across different motion types.  
% We observe particularly strong results in cyclic locomotion, such as walking and dancing, where Lagrangian motion fields effectively capture periodic patterns and maintain consistent quality.  
% Additionally, our method excels at multi-phase sequential actions, such as ``walk-sit-stand," where the supermotion representation naturally segments actions into meaningful units, preserving structural coherence.  

% However, certain limitations emerge in our analysis. For motions involving rapid directional transitions, the uniform-velocity assumption within supermotion segments becomes a constraint. For instance, in the example shown in~\reffig{motion_type}(c), given the prompt ``a person spins around quickly, then abruptly stops and changes direction," we observe a sudden orientation flip between consecutive frames, leading to an unnatural shift in the facing direction.  

% Moreover, our approach struggles with high-frequency localized actions characterized by rapid velocity changes in confined spatial regions.  
% As shown in~\reffig{motion_type}(d), when given the prompt ``a person claps hands two times," the model generates only a single clap.  
% This issue arises from our current clustering strategy, which tends to over-smooth repetitive actions into a single motion segment due to the uniform-velocity assumption within each supermotion.  

We present qualitative visualizations of our method across different motion types in~\reffig{motion_type}. Our approach is particularly effective for dance movements, where the inherent compositional structure, characterized by recognizable step combinations, aligns naturally with supermotion segmentation. Similarly, for multi-phase sequential actions such as ``walk-kneel-stand'', the supermotion representation intuitively segments actions into meaningful units, preserving structural coherence.

For cyclic locomotion, such as walking and running, our method accurately captures and maintains periodic patterns. The supermotion representation naturally preserves the repetitive nature of these motions while ensuring consistency over extended sequences.
Additionally, our approach performs well for large-amplitude motions with stable velocities, such as body twists and bending down. These actions involve smooth velocity transitions, making them particularly suited for Lagrangian motion field modeling. The supermotion representation effectively captures the overall movement trajectory while maintaining natural motion dynamics.

\subsubsection{Visualization of Text-Supermotion Alignment} \label{sec:text_sm_alignment}  
To better understand how our supermotion representation captures semantic meaning from text prompts, we visualize the attention weights between text tokens and motion segments.  
As shown in~\reffig{attention}, for a sequence involving multiple motion types, ``A person waves his arm first, walks forward in a straight line, then turns left," the attention heatmap reveals distinct peaks corresponding to each action: high attention to ``waves" in the initial supermotions, ``walks" in the middle segment, and ``turns" in the final supermotions.  
Similarly, for the sequence ``A person jumps up, and walks forward," we observe strong attention weights between the verb ``jumps" and the initial motion segments (supermotion 0 to 5), followed by high attention to ``walks" in the later segments (supermotion 8 to 14).  
These visualizations demonstrate that our model effectively aligns specific motion segments with their corresponding action words in the text description, highlighting its ability to capture semantic relationships between language and motion.
}

\revise{
\subsubsection{Qualitative Comparison}
Qualitative results for the text-to-motion and music-to-dance tasks are presented in~\reffig{t2m-vis} and~\reffig{m2d-vis}, respectively. Additional examples are provided in the video demonstrations included in the supplemental materials. Compared to baseline methods, our approach generates more diverse and coherent motions over extended durations, demonstrating its effectiveness in long-term motion synthesis.
}

\subsection{User Study}
To evaluate our approach qualitatively, we conducted a pairwise user preference study focused on long-term music-to-dance and text-to-motion tasks. We randomly selected 20 generated motion sequences for each task from the test set and asked 22 volunteers to choose the best one from our method and other comparative methods.

For the long-term music-to-dance generation task, volunteers were instructed to select the dance sequence that exhibited the highest overall visual plausibility and synchronization with the music beats. For the long-term text-to-motion generation task, they were asked to choose the motion sequences that most accurately aligned with the provided long text prompts.

The results of the user study, presented in ~\reffig{user_study}, show that our method demonstrates significant advantages in generation quality compared to other approaches, highlighting its effectiveness and generalization ability across different motion generation tasks.
\subsection{Ablation Study} \label{sec:ablation_study}
\revise{
In this section, we evaluate the key components of our framework, including different coordinate systems, cluster numbers, the proposed coherence loss, and the refinement module, to assess its robustness and generalizability.
}

\textbf{Coordinate Systems.}
Our proposed method functions as a general motion generation pipeline, independent of any specific coordinate system.
To demonstrate the robustness of our approach, we conducted an ablation study to evaluate the impact of different coordinate systems.
In particular, we compared the two most commonly used coordinate systems: Cartesian coordinates (3D) and 6-DOF rotation coordinates (6D).
As shown in \reftab{ablation-m2d} and \reftab{ablation-t2m}, our methods improve the performance of both coordinate systems.
The 6D representation results in slightly improved motion quality compared to the 3D representation, while the 3D representation shows better diversity in both tasks.

\textbf{Cluster Number.} 
We analyze the impact of cluster numbers on performance to provide a more comprehensive ablation study.
Generally, a larger number of clusters reduces the approximation error between the recovered motion and the real motion.
As shown in \reftab{ablation-m2d} and \reftab{ablation-t2m}, smaller cluster numbers can degrade the motion quality and negatively impact both text-motion matching and music-dance alignment. 

\textbf{Coherent Loss.}
We also examined the significance of the coherent loss proposed in our framework.
As indicated in \reftab{ablation-m2d} and \reftab{ablation-t2m}, omitting the coherent loss results in a decline in motion quality.
This underscores the effectiveness of our coherent loss in enhancing motion smoothness and overall quality through the supermotion representation.

\begin{table*}[t]
 \caption{\revise{Ablation Studies on FineDance dataset. ``\#Cls'' denotes the number of clusters. ``Refine.'' indicates whether the Refinement Module is used. ``Params'' shows the number of parameters in millions (M) for the Refinement Module.}}
 \centering 
 \setlength{\tabcolsep}{6pt}
 {
\begin{tabular}{lcccccccccc}
    \toprule
    Methods & \# Cls & $L_{coherent}$ &  \revise{Refine.} & \revise{Params} & BAS$\uparrow$ & FSR $\downarrow$ & $\mathrm{FID}_k\downarrow$ & $\mathrm{FID}_g\downarrow$ & $\mathrm{Div}_k\uparrow$ & $\mathrm{Div}_g\uparrow$ \\
    \toprule
    EDGE (6D)  & -           & -    & -      & - &  0.211 & 20.04\% & 94.34 & 50.38 & 8.13 & 6.45 \\
    Ours (3D)  & 1000        & w/o  & \cmark & 60.78M &  0.201 & 21.17\% & 57.81 & 36.37 & 5.21 & 4.98\\
    Ours (3D)  & 1000        & w    & \cmark & 60.78M &  0.219 & 23.16\% & 53.79 & 33.84 & 5.35 & 5.08\\
    Ours (3D)  & 2000        & w    & \cmark & 60.78M &  0.235 & 24.91\% & 50.27 & 31.63 & 5.63 & 5.27\\
    Ours (6D) & 2000         & w    & \cmark & 60.78M &  0.228 & 12.63\% & 55.49 & 24.93 & 5.46 & 5.82\\
    Ours (6D) & 2000         & w    & \cmark & 41.86M &  0.225 & 12.89\% & 56.82 & 25.74 & 5.83 & 5.76 \\
    Ours (6D) & 2000         & w    & \xmark & - & 0.202 & 36.15\%  & 88.60 & 40.52 & 8.17 & 5.80 \\
    \bottomrule
\end{tabular}\label{tab:ablation-m2d}
}
\end{table*}
\begin{table}[t]
 \caption{Ablation Studies on the HumanML3D dataset.}
 \label{tab:ablation-t2m}
 \centering
 \setlength{\tabcolsep}{2pt}
 {
\begin{tabular}{lcc@{\hskip 5pt}c@{\hskip 5pt}c@{\hskip 5pt}c@{\hskip 5pt}c@{\hskip 5pt}c}
    \toprule
    Methods  & \# Cls & $L_{coherent}$ & R-Pre (top 3)$\uparrow$ & FID$\downarrow$ & MM-Dist$\downarrow$\\
    \toprule
    MDM (6D) &-             & -            & $0.611{\pm.007}$ & $0.544{\pm.044}$ & $5.566{\pm.027}$  \\
    Ours (3D)            & 1000        & w/o & $0.730{\pm.005}$ & $1.760{\pm.030}$ & $3.460{\pm.031}$  \\
    Ours (3D)             & 1000        & w & $0.732{\pm.004}$ & $1.750{\pm.040}$ & $3.455{\pm.029}$  \\
    Ours (3D)             & 2000        & w & $0.737{\pm.004}$ & $1.736{\pm.029}$ & $3.446{\pm.030}$  \\
    Ours (6D)             & 2000        & w & $0.757{\pm.007}$ & $1.138{\pm.027}$ & $3.562{\pm.006}$ \\
    \bottomrule
\end{tabular}
}
\end{table}

\revise{\textbf{Refinement Module.} 
% \input{tab/tab_rec_comp}
% As shown in \reftab{vqvae_comparison}, while learning-based motion compression methods like VQ-VAE and its K-means variant achieve better reconstruction quality, our Lagrangian Motion Fields approach trades some reconstruction accuracy for physical interpretability and direct motion control capabilities.
% To comprehensively analyze the functionality of the refinement module, we conduct a reconstruction comparison between the Proposed supermotion representation and VQ-VAE along with its variant in~\reftab{vqvae_comparison}.
% While learning-based motion compression methods, such as VQ-VAE and its k-means variant, achieve superior reconstruction quality, our Lagrangian Motion Fields approach intentionally trades off some reconstruction fidelity to prioritize physical interpretability and direct motion control capabilities.
% Notably, this limitation is effectively addressed by the Refinement Module without sacrificing interpretability or control. 
To assess the refinement module's impact, we compare performance with and without it in our ablation study.  
As shown in~\reftab{ablation-m2d}, removing refinement (\xmark under ``Refine.")
% \revisetwo{yields higher numerical diversity scores; however, this increase is largely caused by artifacts such as excessive foot sliding and incoherent transitions, which introduce noisy variance rather than meaningful motion differences. In contrast, the refinement module improves temporal coherence and realism, thereby reducing such spurious variance. Although the measured diversity decreases, the refined motions maintain clear perceptual diversity.}
\revisetwo{
degrades motion quality.
Although the ablation variant achieves higher numerical diversity scores, these gains primarily result from artifacts such as excessive foot sliding and incoherent transitions, which introduce noisy variance rather than meaningful motion differences.
In contrast, incorporating the refinement module improves temporal coherence and realism, thereby suppressing spurious variance and substantially enhancing overall quality while preserving perceptual diversity.
}

Moreover, the module is parameter-efficient, as performance remains stable even with fewer parameters. This underscores the supermotion representation’s role in providing structural guidance, enabling efficient quality improvement with minimal overhead.
}

\section{Applications}
The proposed supermotion representation encapsulates the initial pose and defines temporal dynamic information over a specific duration. 
This interpretability makes the representation versatile and applicable across various scenarios. 
Unlike prior methods that primarily emphasize static aspects, such as initial and final poses, while neglecting dynamic attributes like duration, our approach allows explicit control over motion dynamics by incorporating temporal details. 
To demonstrate the effectiveness of the supermotion representation, we present several applications that leverage its interpretability.

\begin{figure}[t]
    \centering
    \includegraphics[width=0.49\textwidth]
    {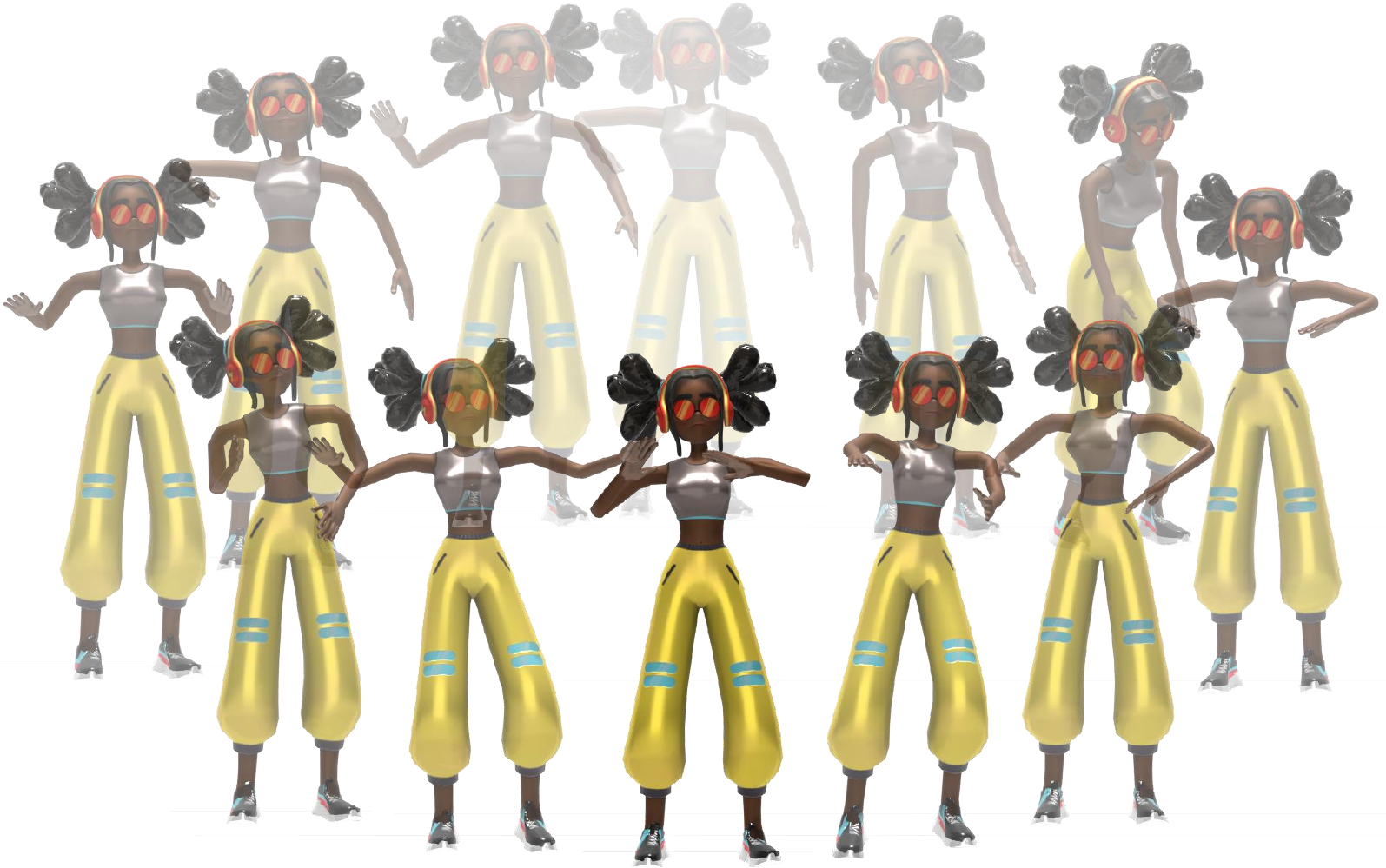}
    % \framebox[\textwidth]{\rule{0pt}{3cm}}
    \caption{Visualization of a generated infinite motion loop with our approach.}
    % \vspace{-4mm}
    \label{fig:loop}
\end{figure}
\textbf{Infinite Motion Looping.}
Generating seamless looping motions requires static and dynamic motion information to ensure that the start and end poses are consistent, with no sudden changes in motion speed. 
Existing motion generation models have not extensively explored the generation of looping dances. Although diffusion model architectures theoretically allow for looping motion generation by aligning the first and last frames, conventional framewise representations lack speed information, often resulting in unnatural transitions at loop points. 
In contrast, our supermotion representation is inherently designed for this task, significantly improving the naturalness and fluidity of looped motion sequences. During the denoising process, we update the final segment of the supermotion as follows:
\begin{equation} \label{eq:app_1}
\mathbf{\hat{sm}_{\tau-1}^{M-1:M}} = \mathbf{\hat{sm}_{\tau-1}^{0:1}},
\end{equation}
where the superscript denotes the supermotion index and the subscript denotes the denoising timestep.
By aligning the first and last supermotions, we ensure both pose alignment and consistent joint velocities between the initial and final ranges.
Our representation inherently includes dynamic information, effectively guaranteeing that motions can transition smoothly from head to tail.
As shown in Fig.~\ref{fig:loop}, the looped dance sequences produced by our approach exhibit a high degree of coherence.

\begin{figure}[t]
    \centering
    \includegraphics[width=0.49\textwidth]{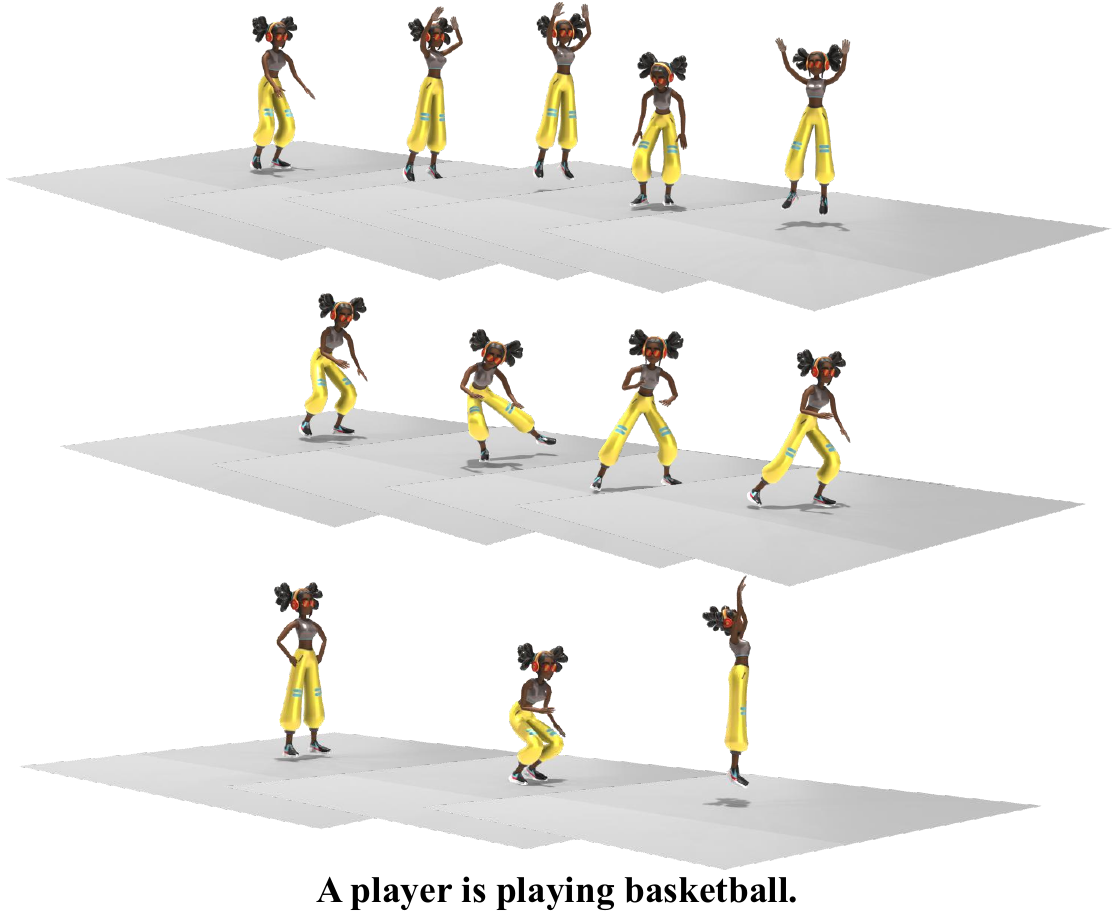}
    \caption{Examples of duration-controlled motion generation. By utilizing our proposed supermotion, the motion with the designated duration for the same text prompt can be generated. Frame numbers from top to bottom are 200, 150, and 100.}
    % \vspace{-4mm}
    \label{fig:duration}
\end{figure}
\textbf{Duration-Controlled Motion Generation.}
In many scenarios, such as virtual storytelling, cinematic animation, or fitness applications, controlling the duration of a motion sequence is essential. 
By leveraging our supermotion representation, which inherently encodes duration information, duration control can be achieved by applying the standard masked denoising technique described by Tseng et. al.~\cite{edge}. 
Given a duration condition $D$, we decompose it into a sequence $[d_0, d_1, \dots, d_{M-1}]$ such that $d_{min} \leq d_i \leq d_{max}$ and $\sum_{s=0}^{M-1} d_{s} = D$. During denoising, we update the duration component of the supermotion as follows:
\begin{equation} \label{eq:app_2}
\mathbf{\hat{sm}_{\tau-1}} = [\mathbf{\hat{x}}_{\tau-1}, \mathbf{\hat{v}}_{\tau-1}, q({d_s}, \tau-1)],
\end{equation}
where the duration part of the supermotion is replaced with a noisy duration condition at each denoising step. 
This approach enables the generation of motion sequences with controllable duration.
As illustrated in Fig.~\ref{fig:duration}, for a given text prompt such as ``A player is playing basketball'', our method can produce variations in the motion sequence’s length while maintaining consistency and realism in the character’s actions.

\section{Discussion and Conclusion}\label{sec:conclusion}

In this paper, we introduced Lagrangian Motion Fields, a novel approach for long-term 3D human motion generation. Our method simplifies temporal representation by producing supermotions, which enhance computational efficiency while maintaining interpretability. Extensive experiments demonstrate that our approach surpasses state-of-the-art methods in tasks such as music-to-dance and text-to-motion generation.

\begin{figure}[t]
   \centering
   \includegraphics[width=\linewidth]{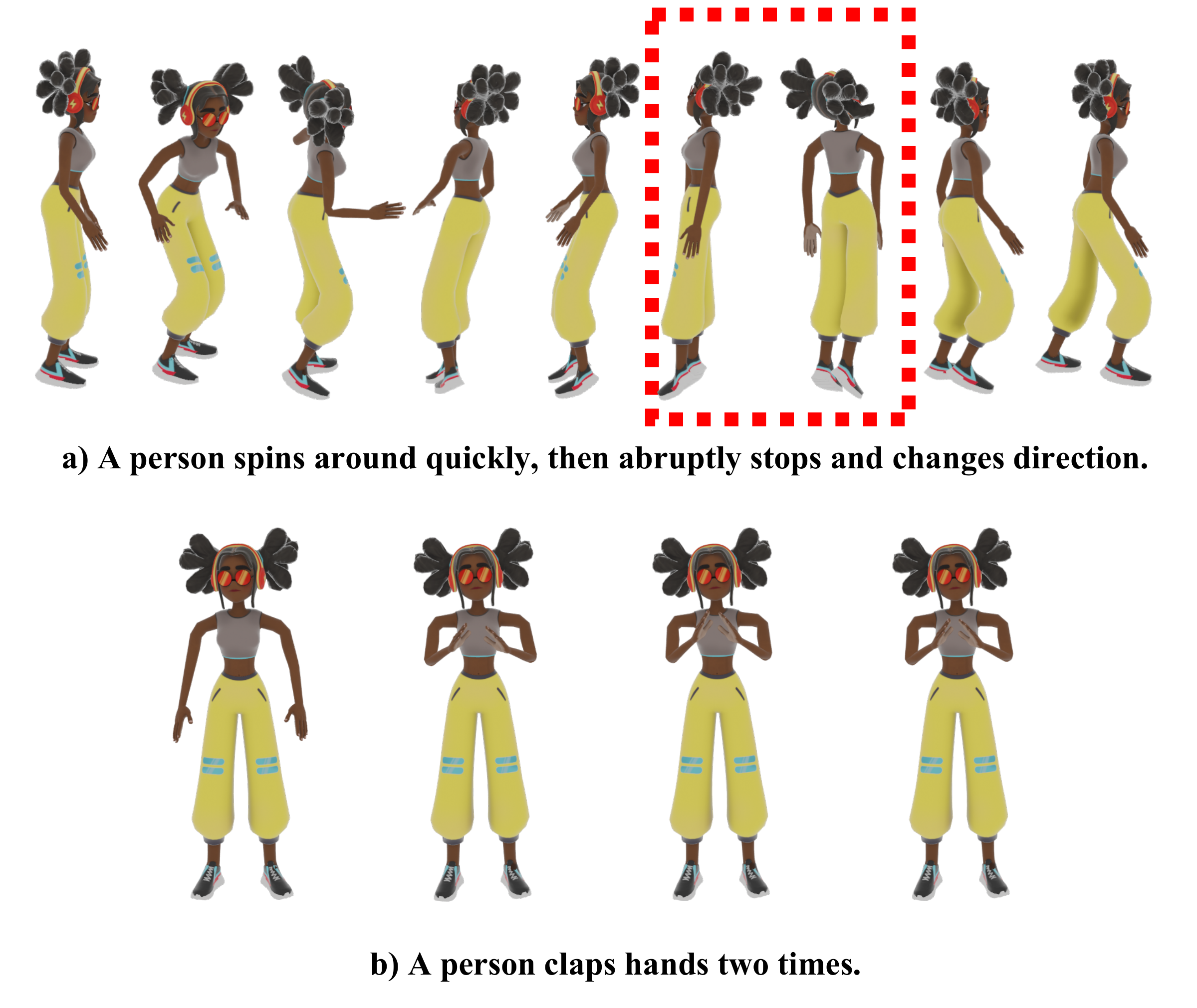}
   \vspace{-8mm}\caption{\revise{Limitations of our method:  (a) Rapid directional changes, leading to frame-to-frame discontinuities.  
(b) High-frequency repetitive actions, where distinct patterns are merged.  
}}
   \label{fig:motion_limit}
    
\end{figure}
\revise{While our method offers significant advantages, abstract representations like supermotions, similar to superpixels and supervoxels, can lead to oversmoothing in motion sequences, slightly affecting realism. Although the proposed lightweight refinement module enhances motion details, it struggles with rapid directional changes and high-frequency localized actions, where abrupt transitions and fine-grained motion nuances are not fully preserved, as shown in Fig.~\ref{fig:motion_limit}.  
To address these limitations, we plan to explore more advanced refinement techniques and adaptive mechanisms to better capture intricate motion dynamics. Furthermore, we aim to extend our approach to real-time motion generation and interactive virtual environments, further validating its versatility and robustness.
}

\bibliographystyle{IEEEtran}
\bibliography{main}

% Ensure all floats are processed
% \FloatBarrier

% Clear the page to ensure the figure is on a new page
% \clearpage

% \begin{appendices}

% %\section*{Figure Pages}
% % \input{fig/fig_loop}
% % \input{fig/fig_duration}
% \clearpage
% % \input{fig/fig_m2d-visualization}
% % \input{fig/fig_t2m-visualization}

% \end{appendices}

\begin{IEEEbiography}[{\includegraphics[width=1in,height=1.25in,clip,keepaspectratio]{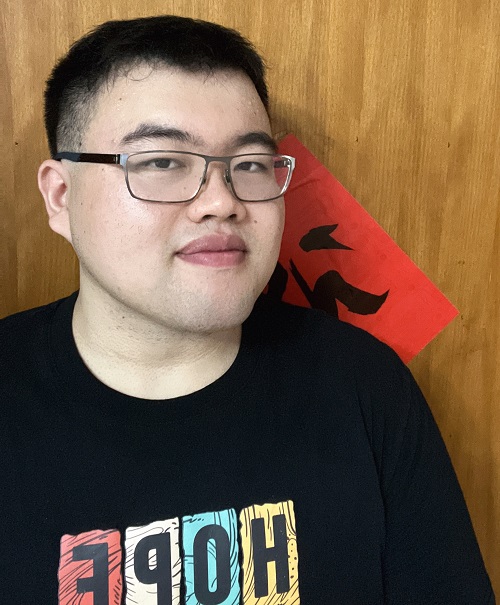}}]{Yifei Yang} is currently pursuing a Ph.D. degree in the School of Computing and Information Systems at Singapore Management University. He obtained his B.Sc. degree from East China Normal University in 2018 and his M.Sc. degree from Shandong University in 2021. His research interests include computer vision and generative models.
\end{IEEEbiography}

\begin{IEEEbiography}[{\includegraphics[width=1in,height=1.25in,clip,keepaspectratio]{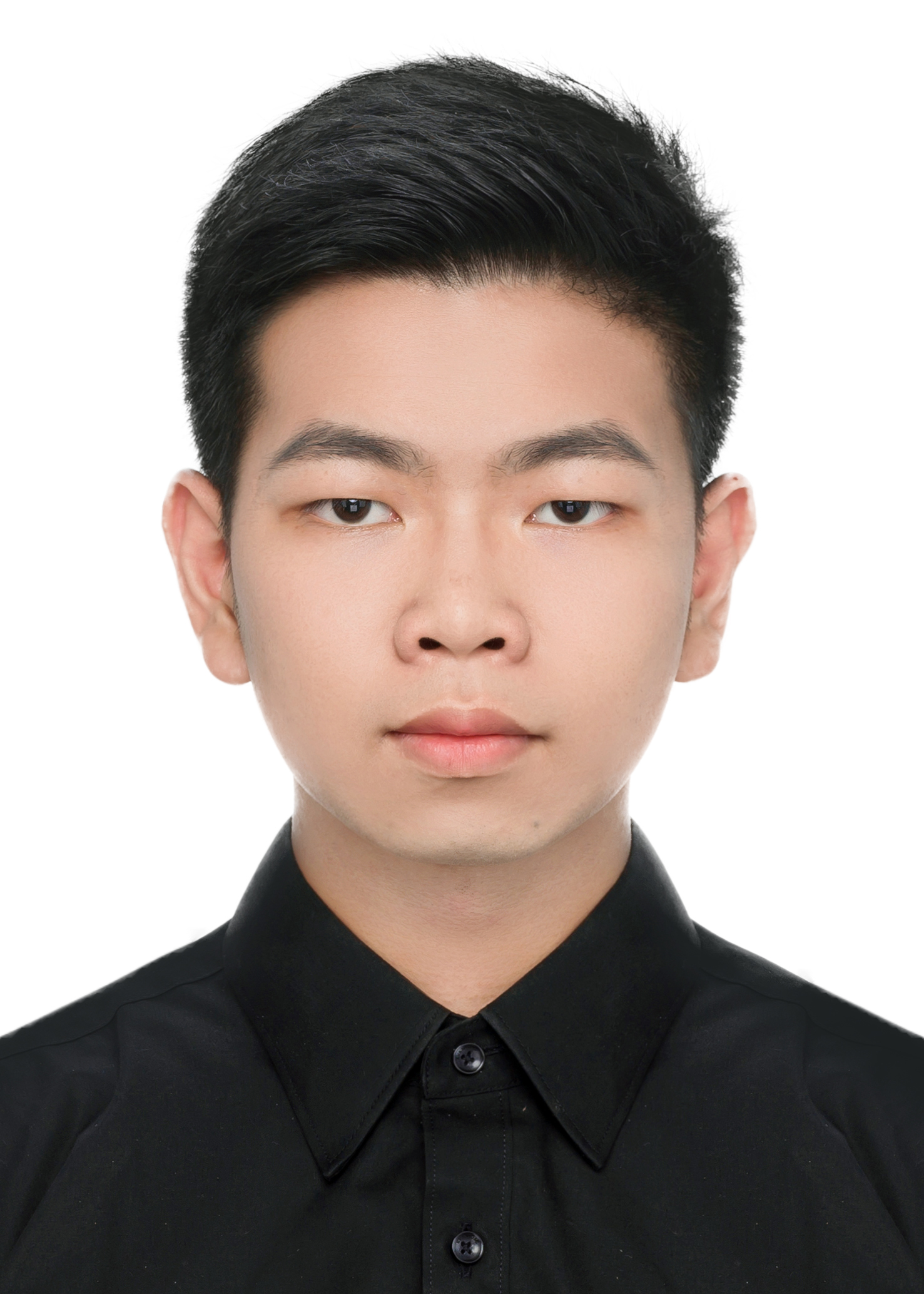}}]{Zikai Huang} is currently pursuing a Ph.D. degree in the School of Computer Science and Engineering, South China University of Technology. His research interests include computer vision, computer graphics and multimodal learning.
\end{IEEEbiography}

\begin{IEEEbiography}[{\includegraphics[width=1in,height=1.25in,clip,keepaspectratio]{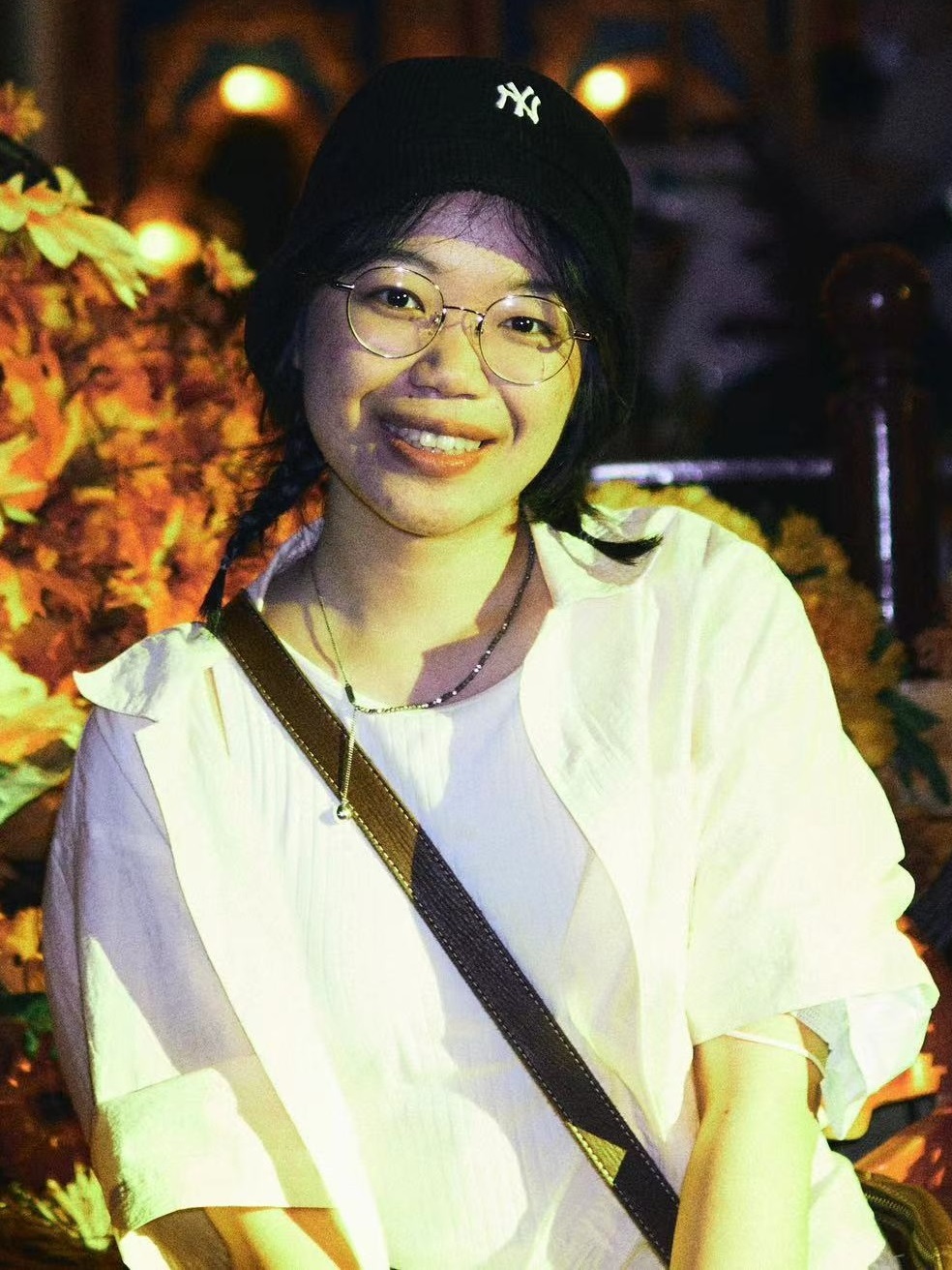}}]{Chenshu Xu} is currently pursuing a Ph.D. degree with the School of Computing and Information Systems, Singapore Management University, Singapore. Her current research interests include computer vision, image processing, computer graphics, and deep learning.
\end{IEEEbiography}\vspace{-4mm}

\begin{IEEEbiography}[{\includegraphics[width=1in,height=1.25in,clip,keepaspectratio]{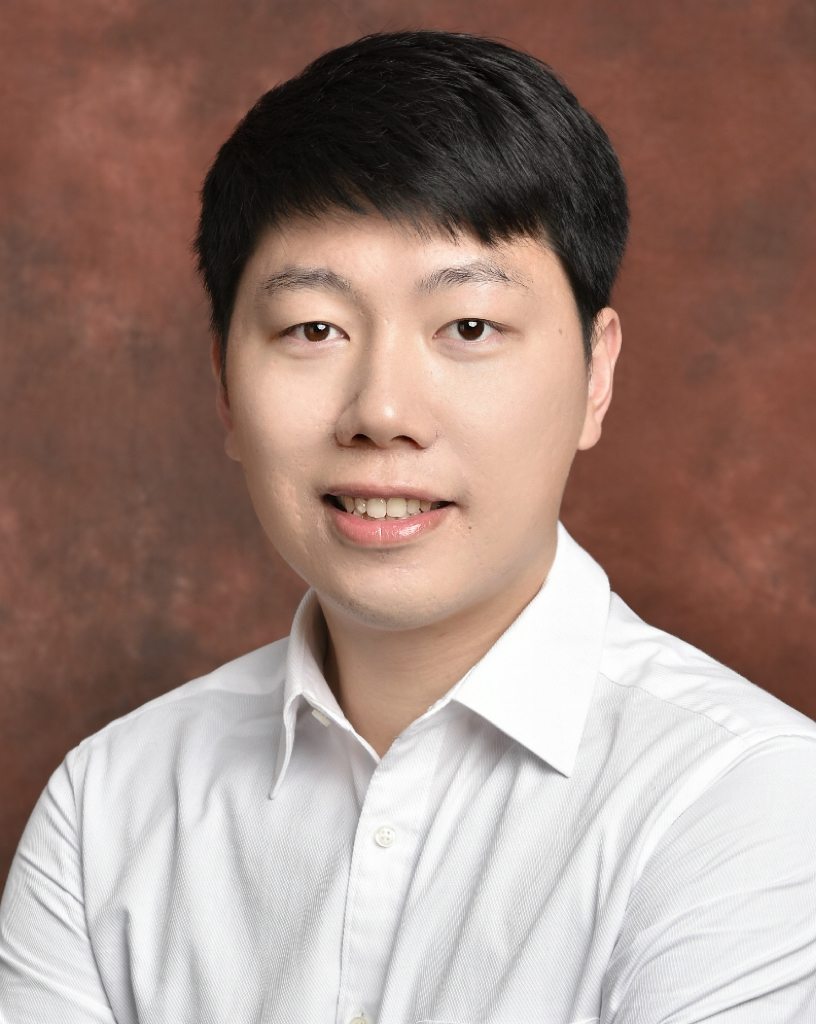}}]{Shengfeng He (Senior Member, IEEE)} is an associate professor in the School of Computing and Information Systems at Singapore Management University. Previously, he was a faculty member at South China University of Technology (2016–2022). He earned his B.Sc. and M.Sc. from Macau University of Science and Technology (2009, 2011) and a Ph.D. from City University of Hong Kong (2015). His research focuses on computer vision and generative models. He has received awards including the Google Research Award, PerCom 2024 Best Paper Award, and the Lee Kong Chian Fellowship. He is a senior IEEE member and a distinguished CCF member. He serves as lead guest editor for IJCV and associate editor for IEEE TNNLS, IEEE TCSVT, Visual Intelligence, and Neurocomputing. He is an area chair/senior PC member for CVPR, NeurIPS, ICLR, ICML, AAAI, IJCAI, BMVC, and the Conference Chair of Pacific Graphics 2026.
\end{IEEEbiography}

\vfill

\end{document}